\documentclass[10pt,twocolumn,letterpaper]{article}

\usepackage[pagenumbers]{wacv} 

\usepackage{graphicx}
\usepackage{amsmath}
\usepackage{amssymb}
\usepackage{booktabs}
\usepackage{nameref}

\usepackage{makecell,algorithm}

\usepackage{adjustbox}

\usepackage{url}  
\usepackage{pifont}
\usepackage{tabularx}
\usepackage[dvipsnames]{xcolor}
\usepackage{pgfplots}
\pgfplotsset{compat=newest} 

\newcommand{\darr}{$\downarrow$}
\newcommand{\uparr}{$\uparrow$}
\newcommand{\cmark}{\ding{51}}%
\newcommand{\xmark}{\ding{55}}%
\newcommand{\textbfit}[1]{\textbf{\textit{#1}}}
\usepackage{xr}
\definecolor{wacvblue}{rgb}{0.21,0.49,0.74}
\usepackage[pagebackref,breaklinks,colorlinks,allcolors=wacvblue]{hyperref}

\usepackage[capitalize]{cleveref}
\crefname{section}{Sec.}{Secs.}
\Crefname{section}{Section}{Sections}
\Crefname{table}{Table}{Tables}
\crefname{table}{Tab.}{Tabs.}

\def\wacvPaperID{181} 
\def\confName{WACV}
\def\confYear{2026}

\begin{document}

\title{\textit{Mem-MLP}: Real-Time 3D Human Motion Generation from Sparse Inputs}

\author{Sinan Mutlu$^{1}$\thanks{Equal Contributions for Main Authorships} \quad Georgios F. Angelis$^{2}$\footnotemark[1] \quad Savas Ozkan$^{1}$\thanks{Senior Authorship} \quad Paul Wisbey$^{1}$  \\ \quad Anastasios Drosou$^2$ \quad Mete Ozay$^{1}$ \\
$^1$ Samsung R\&D Institute UK (SRUK) \\
$^2$ Information Technologies Institute, CERTH \\
s.mutlu@samsung.com, angelisg@iti.gr, savas.ozkan@samsung.com}


\maketitle
\begin{abstract}
   Realistic and smooth full-body tracking is crucial for immersive AR/VR applications. Existing systems primarily track head and hands via Head Mounted Devices (HMDs) and controllers, making the 3D full-body reconstruction incomplete. One potential approach is to generate the full-body motions from sparse inputs collected from limited sensors using a Neural Network (NN) model. In this paper, we propose a novel method based on a multi-layer perceptron (MLP) backbone that is enhanced with residual connections and a novel NN-component called Memory-Block. In particular, Memory-Block represents missing sensor data with trainable code-vectors, which are combined with the sparse signals from previous time instances to improve the temporal consistency. Furthermore, we formulate our solution as a multi-task learning problem, allowing our MLP-backbone to learn robust representations that boost accuracy. Our experiments show that our method outperforms state-of-the-art baselines by substantially reducing prediction errors. Moreover, it achieves $72$ FPS on mobile HMDs that ultimately improves the accuracy-running time tradeoff.
\end{abstract}

\section{Introduction}
\label{sec:intro}

The applications of real-time tracking of 3D human motions are vast and diverse, with significant implications for various fields, including biomechanics, video games, and sports analysis \cite{nogueira2025markerless}. In the context of Virtual Reality (VR) and Augmented Reality (AR) systems, the ability to track 3D human motions in real-time is essential for creating an immersive and interactive experience. By mirroring the user's physical movements in the virtual environment, VR/AR systems can foster a sense of presence and engagement, enabling users to intuitively interact with virtual objects and environments.

\begin{figure}[t!]
\centering
\begin{tikzpicture}
    \begin{axis}[
        width=0.80\columnwidth,
        height=0.6\columnwidth,
        xlabel={\textbf{MPJPE \darr}},
        ylabel={\textbf{Inference Time (FPS) \uparr}},
        legend style={
            font=\footnotesize,
            at={(0.20,0.98)}, 
            anchor=north west, 
            column sep=0.5ex 
        },
        grid=major,
        mark size=4pt, 
        ytick={0, 10, 20, 30, 40, 50, 60, 70, 80}, 
        xtick={2.8, 3.0, 3.2, 3.4, 3.6, 3.8, 4.0, 4.2}, 
        tick label style={font=\bfseries}, 
    ]
    \addplot[
        only marks,
        mark=square,
        color=NavyBlue,
        line width=1.4pt, 
        mark options={solid, thick}, 
    ] coordinates {
        (4.18, 72.0)
    };
    \addlegendentry{AvatarPoser}
    
    \addplot[
        only marks,
        mark=triangle,
        color=red,
        line width=1.4pt,
        mark options={solid, thick},
    ] coordinates {
        (3.93, 26.0)
    };
    \addlegendentry{AGRoL-MLP}
    
    \addplot[
        only marks,
        mark=o,
        color=ForestGreen,
        line width=1.4pt,
        mark options={solid, thick},
    ] coordinates {
        (3.71, 3.5)
    };
    \addlegendentry{AGRoL-Diffusion}
    
    \addplot[
        only marks,
        mark=star,
        color=purple,
        line width=1.4pt,
        mark options={solid, thick},
    ] coordinates {
        (3.08, 72.0)
    };
    \addlegendentry{\textit{Mem-MLP} \textit{(ours)}}
    
    \end{axis}
\end{tikzpicture}
\vspace{-1em}
\caption{On-device comparison of state-of-the-art methods for full-body motion generation. For each method, we report the performance by plotting inference time (FPS) against mean per-joint position error (MPJPE). All the timings are obtained by running the inference on the Quest-3 headset. Note that a lower MPJPE value indicates better accuracy, whereas a higher FPS value indicates better efficiency.}
\vspace{-2em}
\label{ondevicecomparison}
\end{figure}
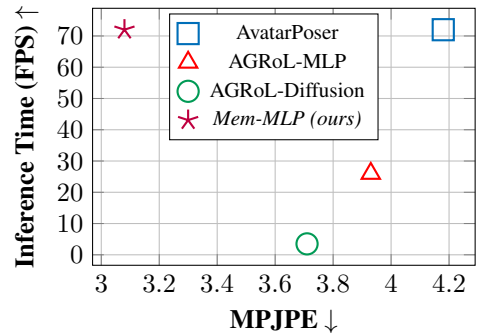

Delivering a fully immersive experience in Virtual Reality (VR) and Augmented Reality (AR) applications relies heavily on tracking full-body movements. However, conventional AR/VR setups are limited by the number of tracking devices, typically consisting of a Head Mounted Display (HMD) and hand controllers equipped with Inertial Measurement Unit (IMUs) devices. As a result, only the movements of head and hands can be tracked and mirrored into the virtual environment, leaving other body parts unobserved. This significant gap in capturing full-body motion hinders the creation of a truly immersive experience, as the virtual environment cannot accurately reflect the user's movements. One potential solution is to attach additional tracking sensors to other body parts, such as the pelvis, legs, or arms \cite{ von2017sparse, jiang2022avatarposer, yang2021real}. However, this approach is often impractical and requires customized setups for different VR/AR scenarios, highlighting the need for a more flexible and efficient solution. To overcome this limitation, a promising approach is to generate 3D motions of the entire human body from the limited data collected by the available tracking sensors. Nevertheless, this task is challenging due to the vast number of possible 3D human motions that can be generated from the sparse input data. Therefore, a robust generation model is necessary to effectively handle these complex scenarios and produces accurate and realistic full-body motions.

\begin{figure}[t]
    \centering
    \includegraphics[width=0.42\textwidth]
    {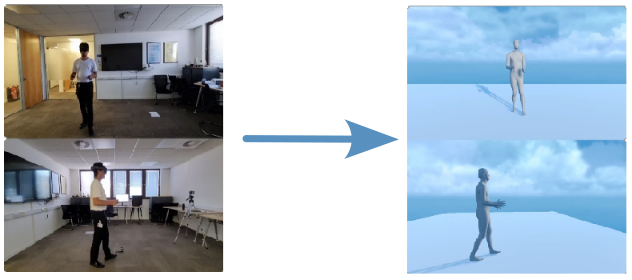}
    \vspace{-0.8em}
    \caption{Given the sparse inputs of body joint representations, our model can accurately generate diverse motions in real-time on a head mounted device.}
    \vspace{-1.5em}
    \label{fig:basic_pipeline}
\end{figure}

In this work, we present a novel method which is designed for generating 3D full-body motions from sparse inputs in real-time as illustrated in Fig.~\ref{fig:basic_pipeline}. Our method builds upon a recently proposed model for 3D human-body motion generation, AGRoL ~\cite{du2023avatars}, which employs a multi-layer perceptron (MLP)-based architecture with residual connections. To improve the accuracy of generated motion sequences, we propose a novel NN-component named \textit{Memory Block} that enhances temporal consistency and reduces errors across various metrics, resulting in more realistic and coherent motion generation. Precisely, the \textit{Memory Block} component employs code-vectors that represent the missing sensor inputs and combines them with the sparse signals from previous time instances. This information is then leveraged in the hidden layers of the neural network (NN)-based backbone. Experimental results on the benchmark AMASS dataset~\cite{mahmood2019amass} demonstrate that the \textit{Memory Block} yields significant performance gains. Moreover, due to its computationally efficient structure, this lightweight component is particularly well-suited for real-time applications.


In contrast to the most state-of-the-art work \cite{jiang2022avatarposer, jiang2024manikin, Aliakbarian_2023_ICCV}, AGRoL \cite{du2023avatars} has shown that adding 3D geometric losses to angle-based losses during training yields only marginal or no performance improvements. One possible explanation is that geometric loss functions (such as distance-based losses for joint positions) and angle-based losses may conflict during the optimization step. Specifically, minimizing joint location errors can lead to suboptimal joint angles, particularly if the model struggles to balance these objectives. To address this issue, we employ a loss weighting mechanism using homoscedastic uncertainty for multi-task learning, which enables our method to leverage both angle-based and distance-based losses. This allows our model to learn more generic and robust representations, ultimately improving the performance of full-body motion generation.   

Lastly, we need to emphasize that our model is considerably lightweight compared to state-of-the-art complex methods that are based on transformers or diffusion models. Our model not only reduces computational overhead but also makes our method highly suitable for real-time applications with high accuracy. We summarize our contributions as follows:

\begin{itemize}
\item We propose a novel NN-based component that represents missing sensor inputs with trainable code-vectors and later utilizes them to add temporal information from previous time instances. It ultimately provides temporal consistency and improves the performance from $22.70$ to $7.00$ for the Jitter metric. The details of our memory-block are presented in Sec.~\ref{sec:backbone}. 

\item A multi-head predictor is introduced for multi-task learning, leveraging the same backbone network to generate both rotation and position predictions. This approach learns more comprehensive representations and delivers an additional improvement from $7.00$ to $6.03$ in the Jitter metric.

\item Our method can be seamlessly integrated into mobile devices, achieving the minimum required runtime efficiency~\cite{MetaHorizon} (i.e., 72 FPS) for interactive applications on head-mounted devices. Additionally, in comparison to the state-of-the-art AvatarPoser~\cite{jiang2022avatarposer}, which also meets the requirements, our method improves the metric accuracy by reducing the position error by $26\%$, as shown in Fig. \ref{ondevicecomparison}. To this end, the trade-off between accuracy and running time is optimized.
\end{itemize}
\vspace{-1pt}
\section{Related Work}
\label{sec:format}
\subsection{Motion Generation with Sparse Inputs}

Motion reconstruction from sparse signal inputs has gained significant attention with the development of the HMDs. Some of the recent research work focused on using six IMU devices, attached to the wrists, lower legs, back and head \cite{von2017sparse}. Similarly \cite{yang2021real} used only four sensors, pelvis, head and hands. In practical extended reality (XR) scenarios, typically only three tracking signals are available: the head and two wrists. Simulation driven pipelines such as \cite{lee2023questenvsim, winkler2022questsim}  leverage physics-based simulation and reinforcement learning to reconstruct plausible motions under sparse inputs. Learning based models such as AvatarPoser \cite{jiang2022avatarposer} address the 3-point setting with transformer architectures, while other approaches cast sparse tracking as a synthesis problem—for example, flow-based models \cite{Aliakbarian_2022_CVPR} and VAE-based methods \cite{Dittadi_2021_ICCV}. Overall, many state-of-the-art methods \cite{jiang2022avatarposer, Dittadi_2021_ICCV, zheng2023realistic, Dai_2024_CVPR, jiang2024egoposer, xia2025envposer} rely on transformer-style designs that are comparatively slower on mobile hardware. In contrast, our model uses an MLP backbone with skip connections (as in \cite{du2023avatars}) delivering competitive accuracy with substantially lower computational cost.

\begin{figure*}[t]
    \centering
    \includegraphics[width=0.85\textwidth]{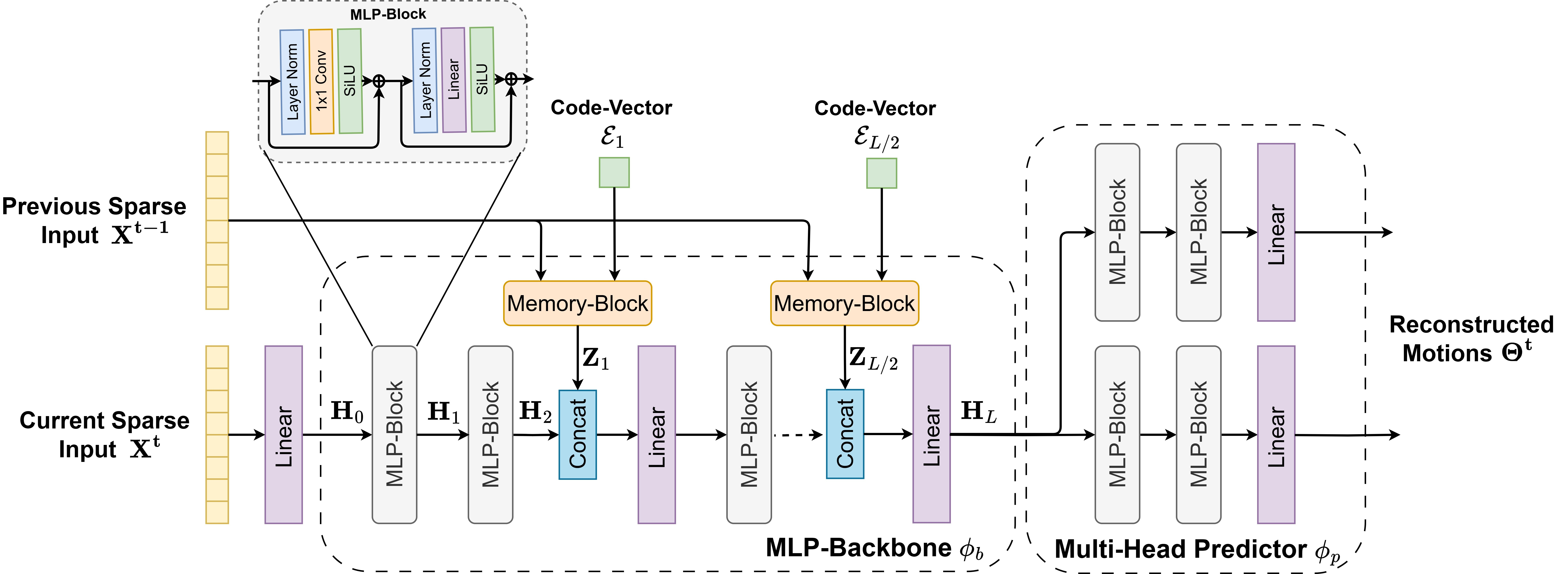}
    \vspace{-0.8em}
    \caption{The architecture of our method. The details of our method is explained in Sec.~\ref{sec:Method}.}
    \vspace{-1.5em}
    \label{fig:arch1}
\end{figure*}

More recent works such as \cite{du2023avatars, feng2024stratified, dong2024realistic, lee2025rewind, tang2024unified} employ diffusion-based architectures. \cite{feng2024stratified} generates separate VQ-VAE \cite{NIPS2017_7a98af17} codebooks for the lower and upper body parts with diffusion process:  first diffusion model generates VQ-VAE code-books using the sparse sensor data, second one generates the VQ-VAE code-books for the lower body using the sparse and upper body code books that are created in the first step. Then these two code-books are given to a decoder to generate all joint angles. Our approach, motivated by \cite{feng2024stratified}, employs code-books as well, hereafter referred to as code-vectors, to integrate human pose and motion priors into the model. This approach allows the generation of smooth and plausible motion sequences using a lightweight autoencoder built from multilayer perceptrons. \cite{dong2024realistic} is based on a U-net shaped architecture which utilizes the Motion Mamba Blocks. Although all of these methods show promising results, they are not feasible for running on mobile devices due to the computational requirement for the \textit{ diffusion process}. In contrast to these methods, our method aligns more closely with \cite{jiang2022avatarposer, Dittadi_2021_ICCV, zheng2023realistic, Dai_2024_CVPR}, generating a motion sequence in a single step. Our design choice not only simplifies the architecture but also enhances the computational efficiency, making it more suitable for real-time applications on mobile devices. 

In a similar domain to 3D human motion generation from sparse inputs, the computer vision and graphics community has focused on generating 3D human motions conditioned on various inputs. These include text \cite{zhang2022motiondiffuse, petrovich2022temos, shafir2023human}, action categories \cite{guo2020action2motion}, incomplete pose sequence completion \cite{duan2021singleshotmotioncompletiontransformer, oreshkin2023motion} and other control signals \cite{starke2022deepphase, jiang2022avatarposer, du2023avatars}. Among these approaches, text-to-motion has emerged as the leading focus in recent years \cite{shafir2023human, zhang2022motiondiffuse, 10096441}, with the aim of learning a shared latent representation between language and motion.

In summary, state-of-the-art works in our domain, as discussed in this section, require either more than three joint inputs to achieve comparable accuracy on the metrics or rely on computationally intensive models, such as transformers. Some of these models also incorporate diffusion samplers, which significantly impact run-time performance on XR devices. In contrast, our proposed method utilizes only three sensor inputs and is built on a simple architecture composed of MLP blocks, complemented by additional lightweight components. Moreover, the simplicity of our method improves its adaptability and feasibility for various XR applications.

\subsection{Motion Dataset}
Modeling human motion has been a persistent challenge in the fields of computer vision and computer graphics. The rise of deep learning has highlighted the growing importance of data for training neural networks to recognize, interpret, and produce human movements. Most commonly existing datasets contain the videos of the human subjects, which are recorded in lab environment, with the 2D or 3D joint key-points \cite{ionescu2013human3}, and each of them employ different body parametrization. For this reason to train our models we used AMASS \cite{mahmood2019amass} dataset which consolidates 15 distinct optical marker-based MOCAP datasets into a single framework with a common parametrization: a skinned vertex based model that accurately represents a wide variety of body shapes in natural human poses, SMPL\cite{loper2023smpl}.

\section{Method}
\label{sec:Method}
\subsection{Problem Formulation}

The function of full-body motion generation $\textbfit{f}$ aims to reconstruct complete full-body motions $\mathbf{\Theta}^\mathbf{t}$ from sparse tracking input signals $\mathbf{X}^{\mathbf{t}}$, which can be represented by
$\mathbf{\Theta}^\mathbf{t}=\textbfit{f}(\mathbf{X}^{\mathbf{t}})$. This function $\textbfit{f}$ typically operates in a time window $\mathbf{t}$ of length $T$, covering time instances from $t-T$ to $t$ to achieve stable motion generation and ensure temporal smoothness. If we define the column-wise concatenation operator by $\mathbf{X}=[X_i]_{i=1}^N \triangleq [X_1, ... , X_{N}]$, within a time window $\mathbf{t}$, the function utilizes the entire set of observed input signals $\mathbf{X}^{\mathbf{t}} = [X^i]^t_{i=t-T} \in \mathbb{R}^ {T \times S}$ to reconstruct the corresponding full-body motions ${\mathbf{\Theta}^\textbf{t} = [\mathbf{\Theta}^i]^t_{i=t-T} \in \mathbb{R}^ {T \times F}}$ for time instances. Here, the dimensions of the input signals and reconstructed motions are denoted by $S$ and $F$, respectively.

\noindent \textbf{Input Signals}: At each time instance $t$, the input signal $X^t$ consists of three sparse signals, collected from the head, left and right hand sensors, and represented by ${X^t = [\mathbf{x}_h^t, \mathbf{x}_l^t, \mathbf{x}_r^t]}$. Each joint signal $\mathbf{x}_j^t$ (where $j \in \{h, l, r\}$) at a time instance $t$ is represented by an axis-angle representation $\mathbf{\psi}_j^t \in \mathbb{R}^3$ and a Cartesian coordinate $\mathbf{p}_j^t \in \mathbb{R}^3$ defined within the global coordinate system. Following previous work~\cite{zhou2019continuity}, the axis-angle representation $\mathbf{\psi}_j^t$ is converted to a 6D representation $\theta_j^t \in \mathbb{R}^6$ by discarding the last row of a rotation matrix. Furthermore, the linear velocity $\mathbf{v}_j^t \in \mathbb{R}^3$ and angular velocity $\mathbf{w}_j^t \in \mathbb{R}^6$ representations are calculated to enforce temporal smoothness, as proposed in~\cite{jiang2022avatarposer}. Consequently, each sensor signal at a time instance $t$ is represented by $\mathbf{x}_j^t = [\mathbf{p}_j^t, \theta_j^t, \mathbf{v}_j^t, \mathbf{w}_j^t]$. The dimension of the input signal $S$ is $54$.

\noindent \textbf{Reconstructed Motion}:
Full-body motions at a time instance $t$ are represented by a tuple $\Theta^t=(\theta_i^t, \mathbf{p}_i^t)_{i=1}^{22} \in \mathbb{R}^ {22 \times 9}$, which comprises the 3D Cartesian coordinates $\mathbf{p}_i^t$ and 6D rotation parameters $\theta_i^t$ for each of the 22 joints of the upper body ($J_{\text{up}}$) and lower body ($J_{\text{lo}}$). To this end, the dimension of the reconstructed full-body motion $F$ is equal to $198$.

\subsection{Our Method}

In our method, the full-body motion generation function $\textbfit{f}$ is implemented by an MLP-based network model $\phi$  parameterized by $\varphi_\phi$. Our architecture is illustrated in Figure~\ref{fig:arch1}. This model is inspired by the architecture proposed in \cite{du2023avatars}, in order to have an efficient and effective full-body motion generation that makes our solution suitable for real-time applications. Our model can be decomposed into two submodels for simplicity: MLP-backbone $\phi_b$ and multi-task predictor $\phi_p$. To this end, it can be represented by $\phi=\phi_p \circ \phi_b$. 

\subsubsection{MLP-backbone:}
\label{sec:backbone}
Our MLP-backbone model $\phi_b$ consists of stacked MLP-blocks and memory-blocks. Details of memory-block will be explained as a separate section. Each MLP-block contains 1D convolution layer, Layer Normalization, SiLU activation, and Linear layer. Intuitively, 1D convolutional layers capture temporal patterns, while linear layers process spatial information from input signals. 

In our implementation, a separate linear layer is initially applied to the sparse signal $\mathbf{X}^\mathbf{t}$ to project it onto a latent representation $\mathbf{H}_0 \in \mathbb{R}^{T \times d}$, where $d$ denotes the dimensionality of the latent space. This latent representation $\mathbf{H}_0$ is then iteratively refined through $L$ numbers of layers of MLP-blocks. The final output of the MLP-backbone is represented by $\mathbf{H}_L \in \mathbb{R}^{T \times d}$, which captures the processed latent information.

\begin{figure}[t]
    \centering
    \includegraphics[width=0.4\textwidth]{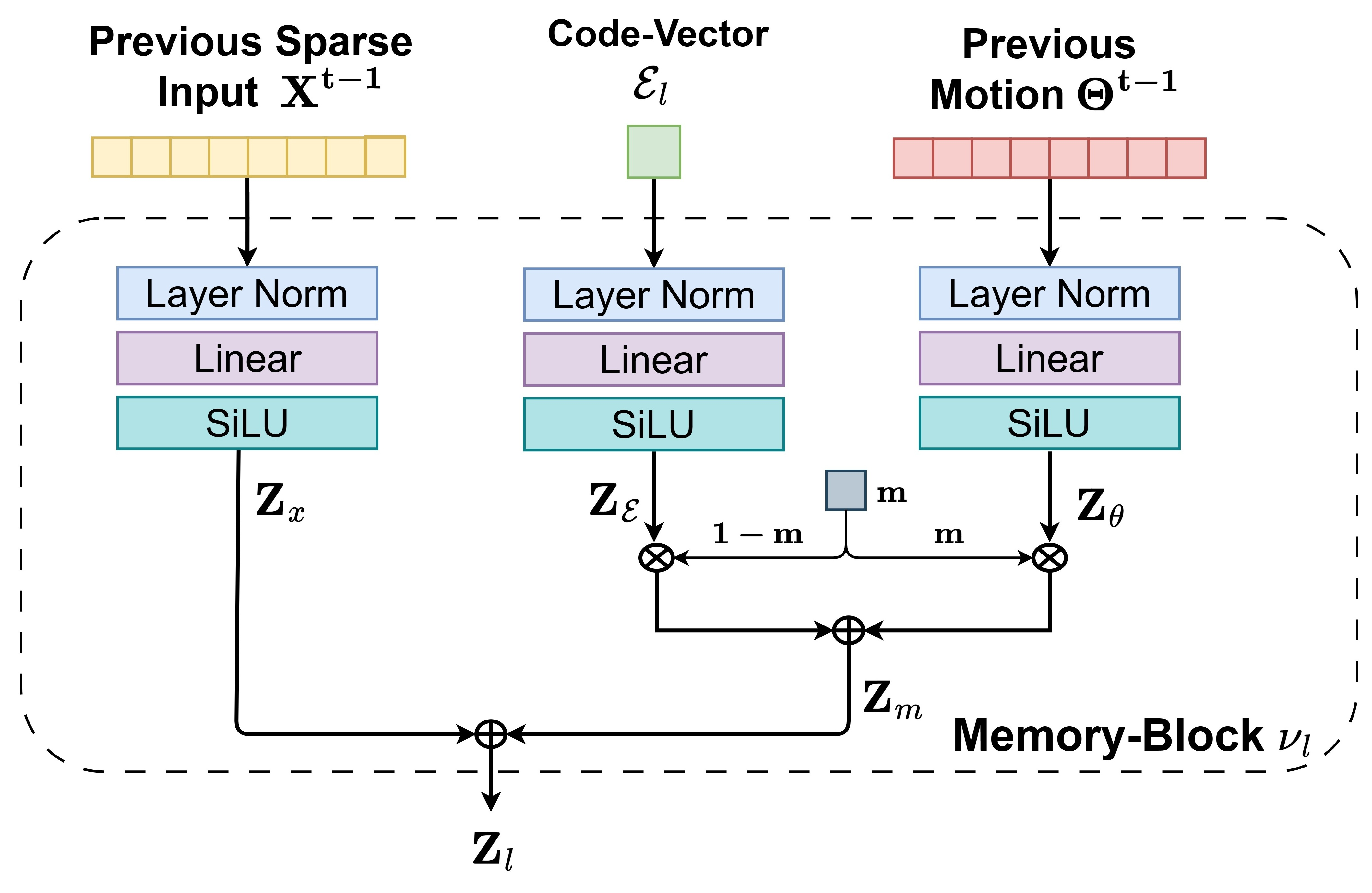}
    \vspace{-0.8em}
    \caption{The architecture of our memory-block component for the $l^{\text{th}}$ layer during training where $l=1,..,L/2$. Its details are explained in Sec.~\ref{sec:backbone}.}
    \vspace{-1.5em}
    \label{fig:memblock}
\end{figure}

\noindent
\textbf{Memory-Block:} 
Generating full-body motions from sparse signals requires the effective use of temporal information to reduce the solution space. However, relying solely on temporal context from sparse signals may be insufficient in edge cases. Increasing the length of the temporal window $T$ can lead to overfitting to past inputs, resulting in overly smoothed reconstructions. To mitigate this, we introduce a neural component called the \textit{Memory-Block}, illustrated in Fig.~\ref{fig:memblock}. It augments sparse observations with learned priors to recover full-body joint dynamics.

In this component, missing signals from joints other than the head and hands are approximated using trainable code-vectors derived from a frozen VQ-VAE model~\cite{feng2024stratified}. Unlike prior use cases our main motivation for exploiting VQ-VAE module is not to generate realistic motions but to encode plausible motion priors conditioned on both sparse signals and corresponding full-body motions at time $t$. Specifically, given sparse features $\mathbf{X}^{t-1}$ and full-body motions $\mathbf{\Theta}^{t-1}$, the encoder produces a latent vector $\mathbf{z}_s = f_{\text{enc}}([\mathbf{X}^{t-1}; \mathbf{\Theta}^{t-1}]), z_s \in \mathbb{R}^{d_{z_{s}}}$, where $d_{z_s} = 256$. This is quantized via a learned codebook $\mathcal{C} = \{ \mathbf{e}_1, \dots, \mathbf{e}_K \} \subset \mathbb{R}^{d_{z_s}}$, where $K=64$ by selecting the closest codebook entry: $\mathcal{E}_l = \mathbf{e}_k \in \mathbb{R}^{256}$ where $k = \arg\min_j \| \mathbf{z}_s - \mathbf{e}_j \|_2$. Each code-vector $\mathcal{E}_l$ thus represents a discrete prior embedding for the $l^{\text{th}}$ memory-block. For more details on the design of VQ-VAE module please refer to the Architectural Design section in the supplementary material. 

During training, the memory-block $\nu_l$ at layer $l$ receives sparse features $\mathbf{X}^{t-1}$, ground-truth motions $\mathbf{\Theta}^{t-1}$, and the corresponding code-vector $\mathcal{E}_l$ to produce a memory-feature: $\mathbf{Z}_l = \nu_l(\mathbf{X}^{t-1}, \mathbf{\Theta}^{t-1}, \mathcal{E}_l)$. Each input is projected to a latent representation using a dedicated MLP (linear layer, layer norm, and SiLU). Let $\mathbf{Z}_x$, $\mathbf{Z}_\theta$, and $\mathbf{Z}_{\mathcal{E}}$ be the resulting latent features. To simulate inference conditions, we blend these features using a random coefficient $\mathbf{m} \in \mathbb{R}^{T\times k}$ is sampled from a uniform distribution on the interval $[0,1]$ to compute $\mathbf{Z}_m = \mathbf{m} * \mathbf{Z}_\theta + (1 - \mathbf{m}) * \mathbf{Z}_{\mathcal{E}}$, where $*$ denotes element-wise multiplication. The final memory feature is then given by $\mathbf{Z}_l = \mathbf{Z}_x + \mathbf{Z}_m$.

At inference time, since ground-truth motions $\mathbf{\Theta}^{t-1}$ are unavailable, we use the predicted full-body motion $\widehat{\mathbf{\Theta}}^{t-1}$ from the previous step. If no previous prediction exists (e.g., at the initial frame), we use a zero tensor as fallback. Thus, the memory-block becomes autoregressive and operates as $\mathbf{Z}_l = \nu_l(\mathbf{X}^{t-1}, \widehat{\mathbf{\Theta}}^{t-1}, \mathcal{E}_l)$, with latent merging and memory-feature computed as before. This autoregressive structure enables the model to leverage temporal continuity while compensating for missing sensor data using learned priors. The resulting memory feature $\mathbf{Z}_l$ is concatenated into the MLP backbone, enhancing intermediate representations across layers.

This design forms an \textbf{autoregressive memory mechanism}, where each memory block utilizes past predictions to generate informative context features for the current frame. The resulting memory-feature $\mathbf{Z}_l$ is concatenated with the hidden representation at the corresponding MLP layer and passed forward through the network. To this end, our novel architecture exploits both information from current and previous time windows and combines them in the latent space. In particular, missing information from joint locations is filled with trainable code-vectors during inference.

\subsubsection{Multi-Task Predictor}
Existing literature predominantly addresses full-body motion generation as a single-task learning problem, focusing on estimating the $6D$ rotations $\theta_i^t, \forall i$ at a time instance $t$ \cite{du2023avatars, feng2024stratified, dong2024realistic, jiang2022avatarposer, Dittadi_2021_ICCV, zheng2023realistic}. In contrast, we adopt a multi-task learning approach, where we not only estimate the rotations but also generate the corresponding global positions $\mathbf{p}_i^t, \forall i$ at a time instance $t$. Our method exploits shared representations to improve the generalization, enabling the model to learn complementary tasks simultaneously and enhancing the overall performance. Ultimately, the prediction of global positions helps to refine the rotational representations by providing spatial context information. On the other hand, accurate rotations lead to more precise positioning, creating a benefit for both tasks. Throughout this process, the MLP-backbone $\phi_b$ is enabled to learn richer features essential for both predictions, since it is responsible for capturing contextual information across the entire input sequence. 
 
 For this purpose, our multi-task predictor 
 $\phi_p$ has two MLP-based branches: one branch generates the joint angle rotations, while the other one generates the joint positions. Each branch has two MLP-blocks and one linear layer. Ultimately, this ensures that the shared representations utilized in both branches improves the precision and robustness of the predictions. 

\subsection{Loss Design}
We utilize multiple loss functions based on angle and position to train the parameters of our network model $\varphi_\phi$. First, an angle loss $\mathcal{L}_{\theta}$ is utilized to minimize the prediction error between ground-truth rotations $\theta^{t}_{i}$ and predicted rotations $\hat{\theta^{t}_{i}}$ for full-body joint locations by
\vspace{-0.5em}
\begin{equation}
\label{rotation_loss}
\begin{split}
\mathcal{L}_{\theta} = \frac{1}{T}  \sum_{t=1}^{T} \sum_{i=1}^{22} \| \theta^{t}_{i} - \hat{\theta}_{i}^t \|_1,
\end{split}
\end{equation}

\noindent
where $T$ indicates the total number of time instances in the training sequence. Furthermore, we introduce a rotational velocity loss $\mathcal{L}_{rv}$ to ensure smooth transitions in temporal motions. It provides an alignment between the predicted angular velocity and the ground truth. This loss is defined as follows:
\vspace{-1.5em}
\begin{equation}
\begin{split}
\mathcal{L}_{rv} = \frac{1}{T-1} \sum_{t=1}^{T-1} \sum_{i=1}^{22} \| (\theta^{t+1}_i - \theta^t_i) - (\hat{\theta}^{t+1}_i - \hat{\theta}^t_i) \|_1.
\end{split}
\end{equation}
For the second branch of our model that computes positions, we include two additional losses to optimize the model parameters. Specifically, a position loss \( \mathcal{L}_{p} \) minimizes the distance between the predicted global positions \( \hat{\mathbf{p}}_i^t \) and the ground truth positions \( \mathbf{p}_i^t \), formulated by:
\vspace{-1.0em}
\begin{equation}
\begin{split}
\mathcal{L}_{p} = \frac{1}{T}  \sum_{t=1}^{T} \sum_{i=1}^{22} \| \mathbf{p}^{t}_{i} - \hat{\mathbf{p}}_{i}^t \|_1.
\end{split}
\end{equation}

To further enhance the smoothness of motion, we introduce a velocity loss  $\mathcal{L}_{fv}$, which encourages the predicted positional velocity to closely match the ground truth velocity. By doing so, we improve the smoothness of the generated motions. The loss is defined by:
\vspace{-0.5em}
\begin{equation}
\begin{split}
\mathcal{L}_{fv} = \frac{1}{T-1} \sum_{t=1}^{T-1} \sum_{i \in J_{\text{lo}}} \| ( \mathbf{p}_{i}^{t+1} - \mathbf{p}^{t}_{i}) - ( \hat{\mathbf{p}}^{t+1}_{i} - \hat{\mathbf{p}}^{t}_{i} ) \|_1
\end{split}
\end{equation}
\vspace{-1.8em}

\noindent
Here, the $\mathcal{L}_{fv}$ loss uses only lower-body joint locations $J_{\text{lo}}$, as the uncertainty associated with reconstructing these joints is typically higher due to the absence of sparse input signals.
The overall objective function $\mathcal{L}$ used for training is defined by:
\vspace{-1.5em}
\begin{equation} 
    \begin{split} 
        \mathcal{L} = & \ \lambda_{\theta} \mathcal{L}_{\theta} 
        + \lambda_{rv} \mathcal{L}_{rv} 
        + \lambda_{p} \mathcal{L}_{p}
        + \lambda_{fv} \mathcal{L}_{fv} 
    \end{split} 
\end{equation}
\vspace{-2.0em}

\noindent
where $\lambda_{\theta}$, $\lambda_{rv}$, $\lambda_{p}$ and $\lambda_{fv}$ are the coefficients that define the contribution of each loss term to the overall objective function. In particular, we utilize an automatic loss weighting mechanism by employing homoscedastic uncertainty \cite{kendall2018multi} to adaptively set these coefficient during training.


\begin{figure}[ht!]
\centering
\begin{tabular}{cc}
\toprule

 \includegraphics[width=3cm,height=3cm]{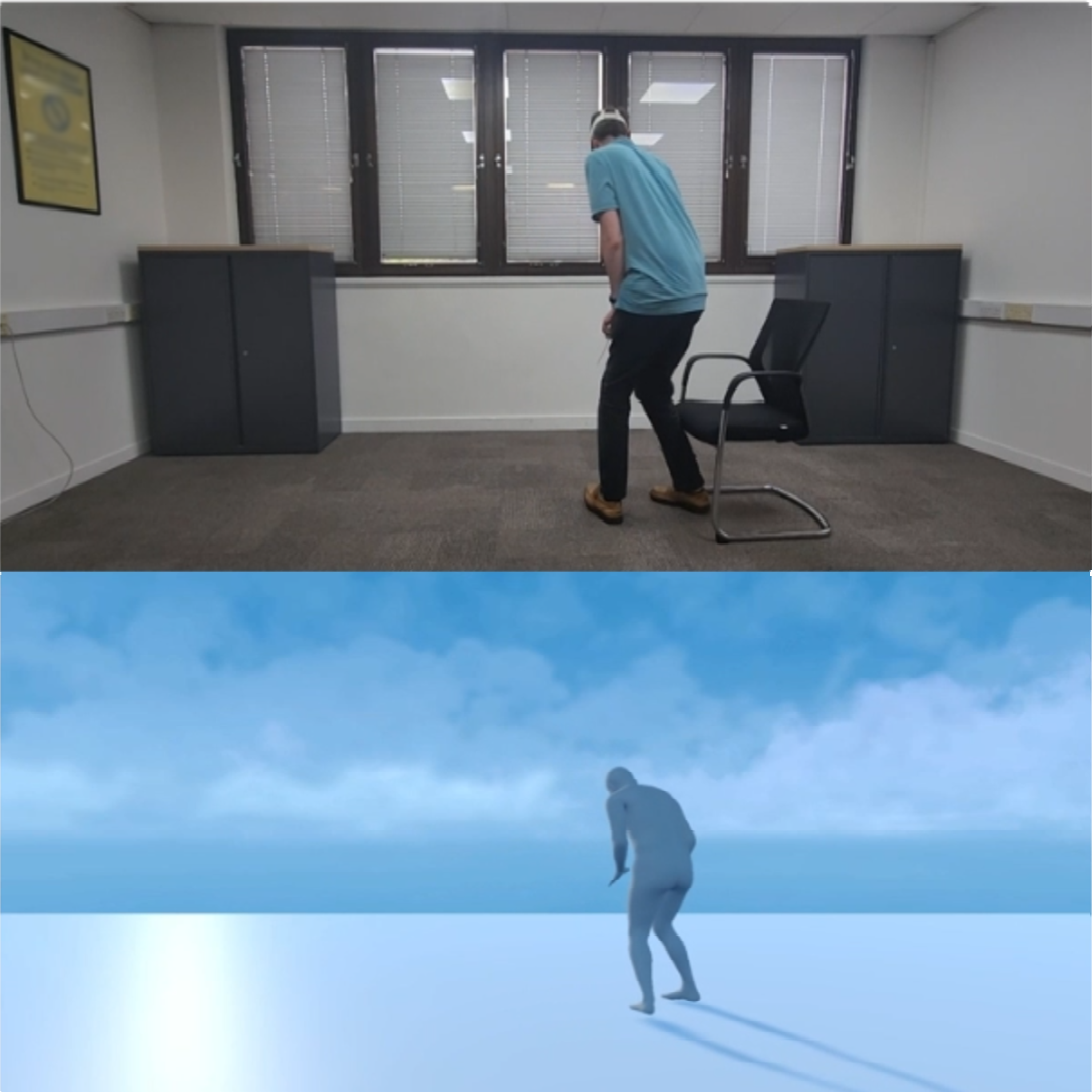}\hspace{-0.1em} &
 \includegraphics[width=3cm,height=3cm]{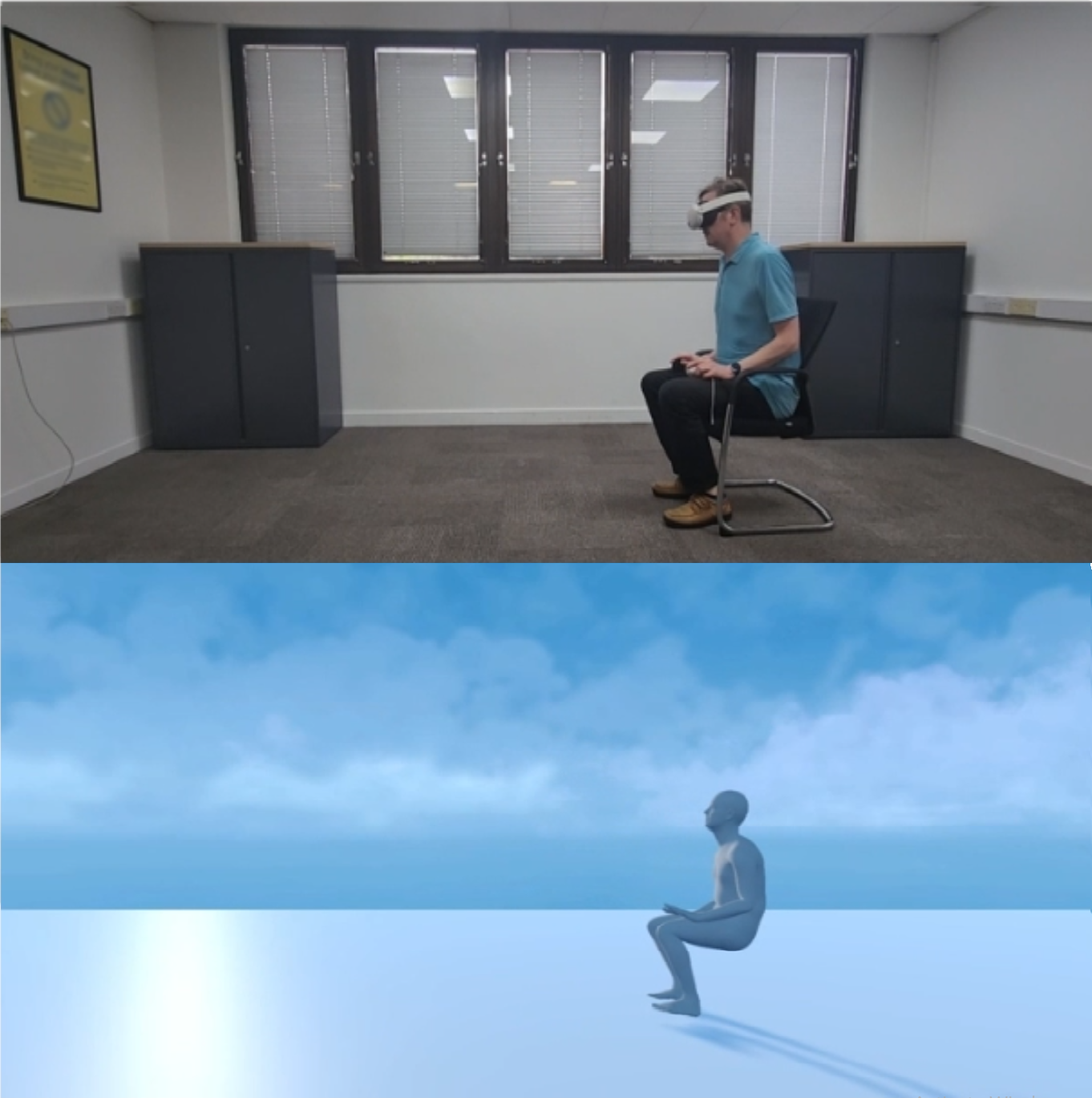} \\
 
 \includegraphics[width=3cm,height=3cm]{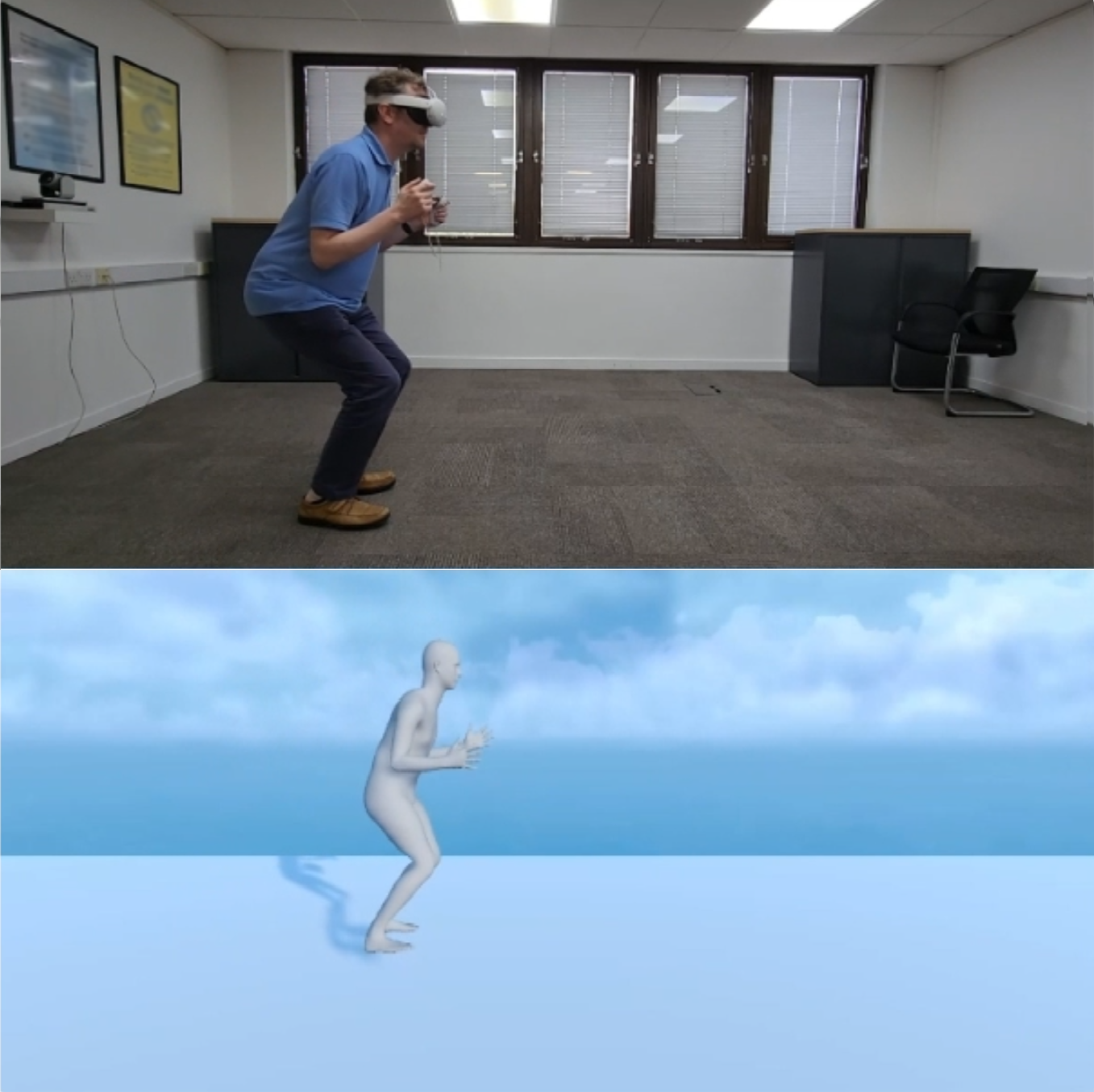}\hspace{-0.1em} &
 \includegraphics[width=3cm,height=3cm]{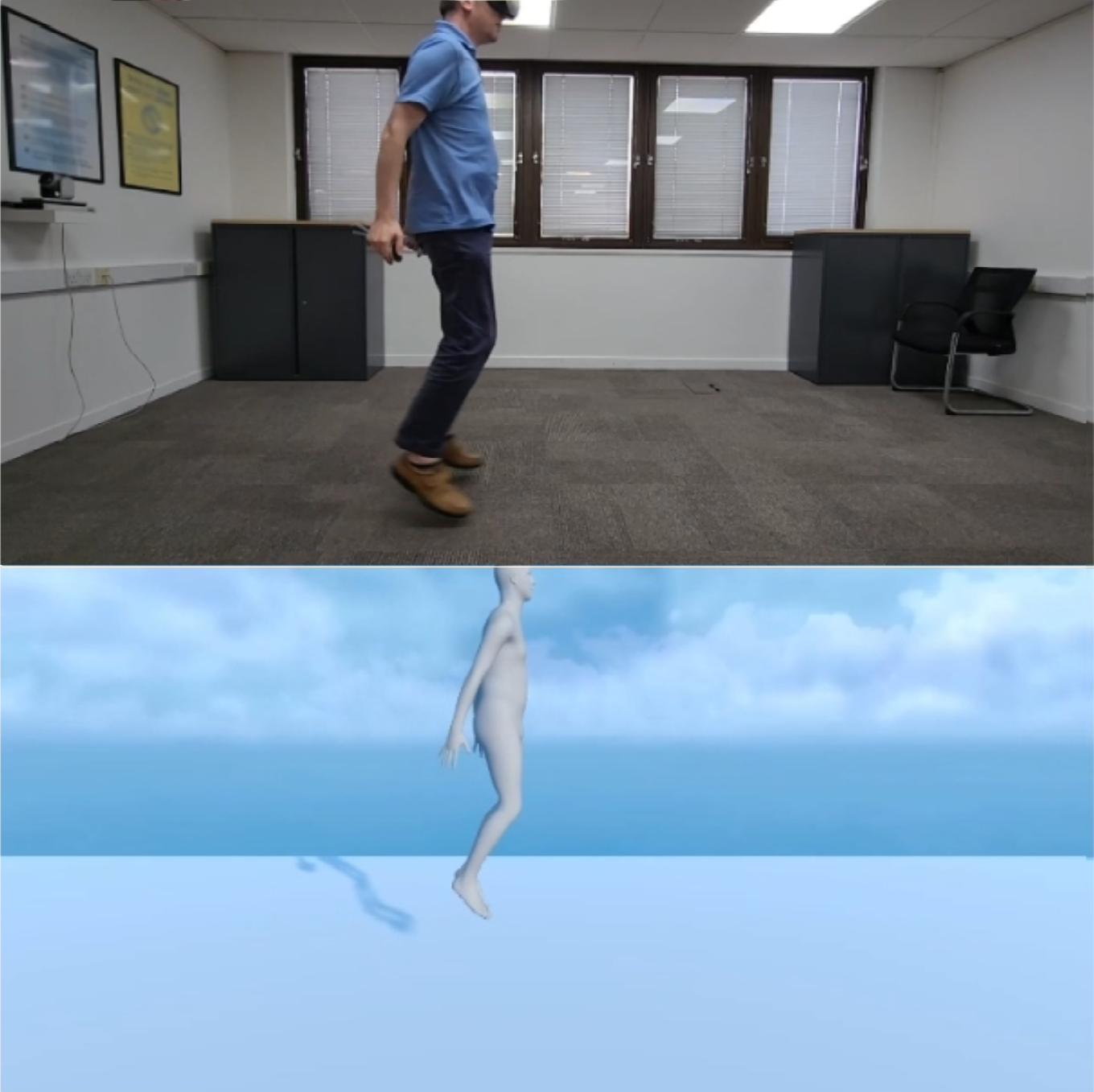} \\
\bottomrule
\end{tabular}
\vspace{-0.8em}
\caption{Visualization of our results on the Quest-3 headset and the hand controllers. Our model is capable to generate diverse motions such as walking, sitting and jumping in real-time.}
\vspace{-1.5em}
\label{fig:visualization}
\end{figure}

\section{Experiments}
\label{sec:exp}
To compare our method with the existing studies, we utilize the AMASS dataset~\cite{mahmood2019amass} as our benchmark for both training and evaluation. In particular, we use this benchmark to accommodate various scenarios.

\noindent
\textbf{Scenario-1 (S1):}
Following the methodology proposed by \cite{jiang2022avatarposer}, we utilize the CMU\cite{CMUDataset}, BMLrub \cite{troje2002decomposing}, and HDM05 \cite{muller2007mocap} subsets by splitting the dataset into 90\% for training and 10\% for testing. 

\noindent
\textbf{Scenario-2 (S2):}
Furthermore, we use the data split introduced in \cite{jiang2022avatarposer, du2023avatars}. In this setup, training is conducted using data from the CMU \cite{CMUDataset}, MPI Limits \cite{akhter2015pose}, Total Capture \cite{trumble2017total},Eyes
Japn \cite{EyesJapan}, KIT \cite{mandery2015kit}, BioMotionLab \cite{troje2002decomposing}, BMLMovi \cite{ghorbani2020movi}, EKUT \cite{mandery2015kit}, ACCAD \cite{ACCAData}, MPI Mosh \cite{loper2014mosh}, SFU \cite{SFUData}, and HDM05 \cite{muller2007mocap} subsets, while evaluation is performed on the HumanEVA \cite{sigal2010humaneva} and Transition \cite{mahmood2019amass} subsets of the AMASS dataset \cite{mahmood2019amass}. This scenario enables us to evaluate the model scalability by testing its performance on motion clips recorded in an unfamiliar environment and also performed by unknown people. As noted in \cite{feng2024stratified}, training/testing partitions are imbalanced in this scenario, resulting a significant differences between test samples and those ones used in training. This can explain the performance drop for our model, similarly reported results in the state-of- the-art works \cite{feng2024stratified, du2023avatars}. 

\noindent
\textbf{Scenario-3 (S3):} All sub-datasets in the AMASS dataset \cite{mahmood2019amass} are divided into training and test sets using the same ratios as in S1. For this specific scenario, we employ the data splits provided by \cite{feng2024stratified}.
\vspace{-1em}
\begin{table*}[!th]
    \centering
    \resizebox{\textwidth}{!}{
    \begin{tabular}{c | ccccccccc}\hline
         \textbf{Method} & \textbf{MPJRE} $\downarrow$ & \textbf{MPJPE} $\downarrow$ & \textbf{MPJVE} $\downarrow$ & \textbf{Hand PE} $\downarrow$ & \textbf{Upper PE} $\downarrow$ & \textbf{Lower PE} $\downarrow$ & \textbf{Root PE} $\downarrow$ & \textbf{Jitter} $\downarrow$  & \textbf{\#FLOPs (G)} $\downarrow$ \\ \hline
         VAR-HMD \cite{Dittadi_2021_ICCV} & $4.11$ & $6.83$ & $37.99$ & -- & -- & -- & -- & -- & -- \\
        AvatarPoser \cite{jiang2022avatarposer} & $3.08$ & $4.18$ & $27.70$ & $2.12$ & $1.81$ & $7.59$ & $3.34$ & $14.49$ & $\underline{0.33}$\\
        AGRoL-Diffusion \cite{du2023avatars} & $2.66$ & $3.71$ & $18.59$ & $1.31$ & $1.55$ & $6.84$ & $3.36$ & $7.26$ & $1.00$\\
        AvatarJLM \cite{zheng2023realistic} & $2.90$ & $3.35$ & $20.79$ & $1.24$ & $1.42$ & $6.14$ & $2.94$ & $8.39$ & $4.64$\\
        BoDiffusion \cite{castillo2023bodiffusion} & $2.70$ & $3.63$ & $14.39$ & $1.32$ & $1.53$ & $7.07$ & -- & $\mathbf{4.90}$ & $0.46$\\
        SAGE Net \cite{feng2024stratified} & $2.53$ & $3.28$ & $20.62$ & $1.18$ & $1.39$ & $6.01$ & $2.95$ & $6.55$ & $4.10$\\
        EgoPoser \cite{jiang2024egoposer} & $3.09$ & $5.24$ & $24.93$ & $4.97$ & $3.79$ & $7.80$ & $3.78$ & $15.37$ & $0.33$\\ 
        MANIKIN-S \cite{jiang2024manikin} & $-$ & $3.36$ & $23.18$ & $\underline{0.02}$ & $1.32$ & $6.72$ & $-$ & $7.95$ &  $0.33$ \\ 
        MANIKIN-LN \cite{jiang2024manikin} & $-$ & $\mathbf{2.73}$ & $\mathbf{13.55}$ & $\mathbf{0.01}$ & $\underline{1.30}$ & $\mathbf{5.13}$ & $-$ & $7.95$ &$4.64$\\ 
        MMD \cite{dong2024realistic} & $\underline{2.30}$ & $3.17$ & $17.32$ & $0.79$ & $\mathbf{1.25}$ & $5.94$ & $2.86$ & $6.52$ &$7.98$\\ 
        HMD-Poser \cite{Dai_2024_CVPR} & $\mathbf{2.28}$ & $3.19$ & $17.47$ & $1.65$ & $1.67$ & $\underline{5.40}$ & $3.02$ & $6.07$ &$0.38$\\  
        \hline
        \hline
        AGRoL-MLP-196 \cite{du2023avatars} & $2.69$ & $3.93$ & $22.85$ & $2.62$ & $1.89$ & $6.88$ & $3.35$ & $13.01$ & $0.88$\\
        \hline \hline
        \textit{Mem-MLP-41} \textit{(ours)} & $2.57$ & $3.08$ & $15.02$ & $0.27$ & $\mathbf{1.25}$ & $5.72$ & $\mathbf{2.85}$ & $6.03$ & $\mathbf{0.25}$\\ 
        \textit{Mem-MLP-60} \textit{(ours)} & $2.56$ & $\underline{3.05}$ & $\underline{14.27}$ & $0.25$ & $1.26$ & $5.67$ & $\underline{2.87}$ & $\underline{5.35}$ & $0.38$\\
        \hline
    \end{tabular}
    }
    \vspace{-0.8em}
    \caption{Comparison with state-of-the-art methods for full-body motion estimation. Results on Scenario-1 (S1) are reported on AMASS dataset for Jitter [m/s\textsuperscript{3}], MPJVE [cm/s], MPJPE [cm], Hand PE [cm], Upper PE [cm], Lower PE [cm], MPJRE [deg] and GFLOPs. The best results are in \textbf{bold}, and the second best results are \underline{underlined}.}
    \label{tab:overall_results}
    \vspace{-1em}
\end{table*}

\subsection{Implementation Details}
Following previous works, we set the time window length $T$ to $41$ to effectively capture the temporal information. This value is also used for the input dimensions of the memory-blocks. Our MLP-backbone consists of $L=8$ stacked MLP-blocks, where the outputs of the memory-blocks are concatenated with the outputs of the even-numbered MLP blocks. In multi-task predictor, each branch has two MLP-blocks followed by a linear layer. Our model is trained for $300$K steps using the AdamW optimizer, starting with an initial learning rate of $3e-4$ that is reduced to $1e-5$ after $225$K steps.

\subsection{Evaluation Metrics}

For the evaluation of our method, we utilize the standard metrics used in the literature~\cite{jiang2022avatarposer, du2023avatars}:

\noindent
\textbf{Rotation-related metric:} The mean per joint rotation error [degrees] (\textit{MPJRE}) measures the average relative rotation
error for all joints.

\noindent
\textbf{Velocity-related metrics:} These metrics consist of the mean per joint velocity error [cm/s] (\textit{MPJVE}) and Jitter. \textit{MPJVE} calculates the average velocity error across all joint positions. Jitter serves as a measure of how smooth the motion is by computing the rate of change in acceleration over time for all body joints in global space, \(10^2m/s^3\).

\noindent
\textbf{Position-related metrics}: The mean per joint position error [cm] captures the average deviation in position across all joint locations. \textit{Root PE} specifically measures the positional error of the root joint, while \textit{Hand PE} computes the average error for both hands. Additionally, \textit{Upper PE} focuses on the joints in the upper body, and \textit{Lower PE} targets those in the lower body.

\subsection{Evaluation Results}
\label{eval_results}

\begin{figure*}[!htb]
\begin{minipage}[b]{0.95\textwidth}
\centering
\begin{tabular}{cc}
\includegraphics[width=0.9\textwidth]{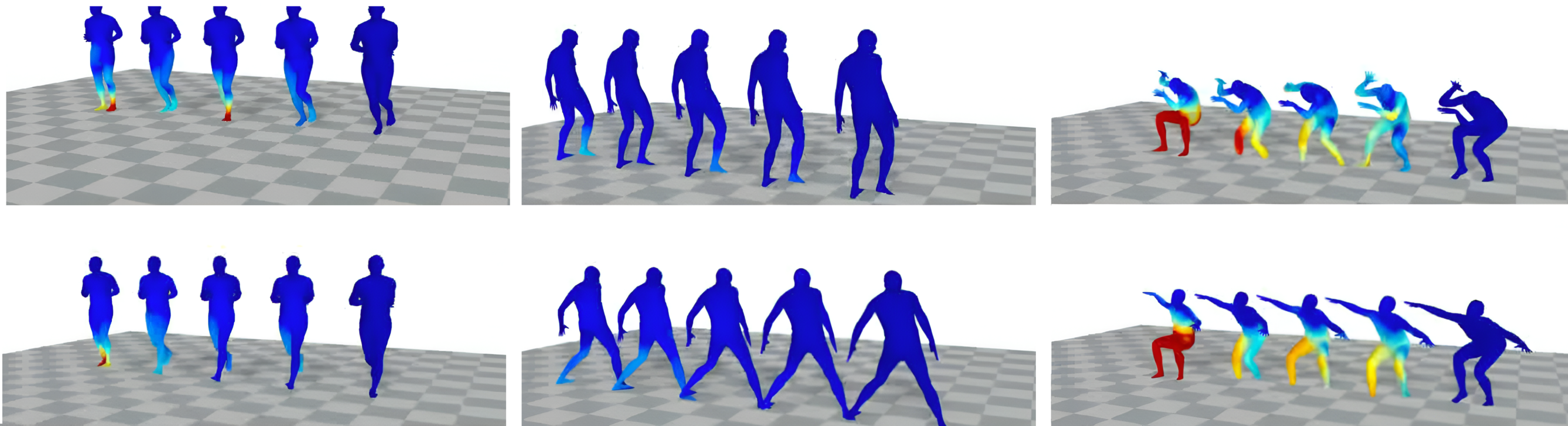}\hspace{-0.01em}  & 
\includegraphics[width=0.05\textwidth]{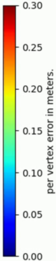} \\
\end{tabular}
\vspace{-1.0em}
\caption{Motion reconstruction results of \textit{running}, \textit{sitestep} and \textit{pterosaur} motions in Scenario-1. Heat-map errors are plotted for the compared methods: from left to right, AvatarJLM \cite{zheng2023realistic}, AGRoL-Diffusion \cite{du2023avatars}, SAGE-Net \cite{feng2024stratified}, ours (\textit{Mem-MLP-41}) and the ground-truth.}
\vspace{-1.0em}
\label{Fig:poses_3}
\end{minipage}
\end{figure*}

\noindent
\textbf{Evaluation Results on Scenario-1 (S1):}
Table \ref{tab:overall_results} presents a detailed comparison of our Mem-MLP model with several baseline models for full-body motion generation.  In particular, we evaluated two variants of our \textit{Mem-MLP} model, each configured with different window lengths of $T=41$ and $T=60$. Furthermore, we include the results of our base model, AGRoL-MLP \cite{du2023avatars}, which uses a time window of $T=196$, Fig-\ref{Fig:poses_3}.

The results demonstrate that both versions achieve comparable performance across most metrics; however, the Jitter metric reveals that longer input lengths yield smoother motion reconstruction. Furthermore, these two models achieve competitive results to the best performing ones in the standard metrics, surpassed only by the resource-intensive model BoDiffusion \cite{castillo2023bodiffusion}. Although MDM \cite{dong2024realistic} shows strong accuracy in the S1 dataset, its performance is not reported for other scenarios. 

In Table \ref{tab:overall_results}, FLOPs are also reported for all models. \textit{Mem-MLP-41} has the lowest FLOPs (0.25 G) compared to other models, demonstrating a clear advantage in terms of performance per computational cost. On the other hand, while \textit{Mem-MLP-60} requires slightly higher computational cost (0.38 G), it achieves better accuracy in several metrics. To this end, while \textit{Mem-MLP-41} offers a faster alternative, \textit{Mem-MLP-60} provides superior accuracy, making it ideal for use cases that prioritize precision over speed. Note that the metric results for EgoPoser \cite{jiang2024egoposer} are computed by running their model checkpoint on our S1 dataset, it was not reported in \cite{jiang2024egoposer}, for direct comparison and performance evaluation.

\begin{table}[t!]
\centering
\resizebox{\columnwidth}{!}{%
\begin{tabular}{lcccc}
\hline
\textbf{Method} & \textbf{MPJRE} \darr & \textbf{MPJPE} \darr & \textbf{MPJVE} \darr & \textbf{Jitter} \darr \\
\hline
VAE-HMD \cite{Dittadi_2021_ICCV} & -- & $7.45$ & -- & -- \\
AvatarPoser \cite{jiang2022avatarposer} & $4.70$ & $6.38$ & $34.05$ & $10.21$ \\
AGRoL-Diffusion \cite{du2023avatars} & $\underline{4.30}$ & $6.17 $& $24.40$ & $8.32$ \\
AvatarJLM \cite{zheng2023realistic} & $ \underline{4.30}$ & $\mathbf{4.93}$ & $26.17$ & $7.19$ \\
BoDiffusion \cite{castillo2023bodiffusion} & $ 4.53$ & $5.78$ & $\mathbf{21.37}$ & $\mathbf{3.50}$ \\
HMD-Poser \cite{Dai_2024_CVPR} & $ 4.27 $ &  $5.44 $ & $30.15$ & $\mathbf{5.62}$ \\
SAGE Net \cite{feng2024stratified} & $4.62$ & $5.86$ & $33.54$ & $\underline{7.13}$ \\
\hline
AGRoL-MLP-196 \cite{du2023avatars} & $ 4.33 $ & $6.66$ & $33.58$ & $21.74$ \\
\hline \hline
\textit{Mem-MLP-41} \textit{(ours)} & $4.33$ & $5.24$ & $23.70$ & $7.23$ \\
\textit{Mem-MLP-60} \textit{(ours)} & $\mathbf{4.02}$ & $\underline{5.13}$ & $\underline{22.60}$ & $6.06$ \\
\hline
\end{tabular}}
\vspace{-0.8em}
\caption{Comparison with state-of-the-art methods for full-body motion estimation. Results on Scenario-2 (S2) are reported on AMASS dataset for MPJVE [cm/s], MPJPE [cm],  MPJRE [deg] and Jitter [m/s\textsuperscript{3}]. The best results are in \textbf{bold},  and the second best results are \underline{underlined}.}
\vspace{-1.5em}
\label{table:comparison_s2}
\end{table}


\begin{table*}[t!]
    \centering
    \resizebox{\textwidth}{!}{%
    \begin{tabular}{lcccccccc}
        \hline
        \textbf{Method} & \textbf{MPJRE} \darr & \textbf{MPJPE} \darr & \textbf{MPJVE} \darr & \textbf{Hand PE} \darr & \textbf{Upper PE} \darr & \textbf{Lower PE} \darr & \textbf{Root PE} \darr & \textbf{Jitter} \darr \\
        \hline
        Avatarposer \cite{jiang2022avatarposer}  & $2.72$ & $3.37$ &$ 21.00$  & $2.12$ & $1.63$ & $5.87$ & $2.90$ & $10.24$  \\
        AGRoL-Diffusion \cite{du2023avatars} & $2.83$ & $3.80$ & $17.76$  & $1.62$ & $1.66$ & $6.90$ & $3.53$ & $10.08$  \\
        AvatarJLM \cite{zheng2023realistic}       & $3.14$ & $3.39$ & $\underline{15.75}$  & $\underline{0.69}$ & $1.48$ & $6.13$ & $3.04$ & $5.33$   \\
        SAGE Net \cite{feng2024stratified} & $\underline{2.41}$ & $\underline{2.95}$ & $16.94$  & $1.15$ & $\underline{1.28}$ & $\underline{5.37}$ & $\underline{2.74}$ &$\underline{5.27}$   \\
        \hline
        AGRoL-MLP-196 \cite{du2023avatars}  & $3.09$ & $4.31$ & $109.29$ & $1.79$ & $1.80$ & $7.95$ & $3.86$ & $12.18$ \\
        \hline \hline
        \textit{Mem-MLP (ours)} & $\mathbf{2.27}$ & $\mathbf{2.54}$ & $\mathbf{11.38}$  & $\mathbf{0.23}$ & $\mathbf{1.08}$ & $\mathbf{4.66}$ & $\mathbf{2.47}$ & $\mathbf{5.10}$\\
        \hline
    \end{tabular}}
    \vspace{-0.8em}
    \caption{Comparison with state-of-the-art methods for full-body motion estimation. Results on Scenario-3 (S3) are reported on AMASS dataset for Jitter [m/s\textsuperscript{3}], MPJVE [cm/s], MPJPE [cm], Hand PE [cm], Upper PE [cm], Lower PE [cm] and MPJRE [deg]. The best results are in \textbf{bold}, and the  second best results are \underline{underlined}.}
    \label{table:evaluation_s3}
    \vspace{-1em}
\end{table*}

\noindent
\textbf{Evaluation Results on Scenario-2 (S2):}
Table \ref{table:comparison_s2} presents the experimental results for Scenario-2. Our method demonstrates competitive performance in S2, achieving superior results across all metrics compared to alternative models such as AvatarPoser \cite{jiang2022avatarposer} and AGRoL-MLP\cite{du2023avatars} that can have real-time inference. 
However, the observed degradation in accuracy can likely be attributed to the limited diversity within the overall data distribution of the S2 training dataset. Specifically, certain motion types such as crawling, kicking, jumping, twisting, and dancing are either not represented or underrepresented in the Transition \cite{mahmood2019amass} dataset. This lack of motion variety may hinder a thorough assessment of the models’ scalability, potentially leading to less reliable evaluation outcomes. 

\noindent
\textbf{Evaluation Results on Scenario-3 (S3):}
Table \ref{table:evaluation_s3} presents the evaluation results for Scenario-S3. Our \textit{Mem-MLP} model demonstrates superior performance across all metrics, further validating its superiority. Notably, our model achieves outstanding results in MPJVE, outperforming even state-of-the-art methods such as AvatarJLM \cite{zheng2023realistic} and SAGE Net \cite{feng2024stratified}. This highlights the capability of our model to generate smoother and more natural motion estimates. Note that some of the models listed in Table \ref{eval_results} are absent from Table \ref{table:evaluation_s3} as they were not made available in the original publication.

\noindent
\textbf{Accuracy vs Inference Time:}
We compare MPJRE in Scenario-1 (S1) against on-device inference time measured in frames per second (FPS) in Fig.~\ref{ondevicecomparison}. Specifically, various baseline models are deployed on the Quest-3 device. Notably, our model achieves the maximum FPS value of $72$, the minimum required runtime efficiency limit for ensuring smooth and seamless interaction. Avatarposer \cite{jiang2022avatarposer} is also suitable for real-time applications, as it achieves similar FPS performance as our model. However, its predictive accuracy is significantly low, with notably high error rates, making it less reliable for tasks requiring precise motion estimation. Additionally, AGRoL-Diffusion \cite{du2023avatars} achieves only $3.5$ FPS, rendering it unsuitable for real-time deployment due to its limited computational efficiency. In contrast, running at $29$ FPS on the device, AGRoL-MLP \cite{du2023avatars} is a more feasible option for deployment, but it is still well under the minimum required rendering efficiency due to its larger time window input. Notably, our model stands out as one of the best top-performing models, uniquely combining real-time runtime capabilities on XR headsets with high accuracy levels. We also visualize \textit{Mem-MLP} results on the Quest-3 headset in Fig.~\ref{fig:visualization} for diverse motion cases.

\subsection{Ablation Study}
\label{ablations}

\noindent
\textbf{Impact of Multi-Head Predictor and Memory Block:} 
In this section, we present an ablation study to analyze the impact of multi-head predictor and memory block for our \textit{Mem-MLP} model under the Scenario-1 (S1). For this purpose, we train a MLP model that corresponds to the AGRoL-MLP-41~\cite{du2023avatars} with a smaller time window (i.e., 41) to reduce the computational overhead compared to its original version. We evaluate the impact of Multi-Head Predictor and Memory-Block on this base model. As shown in Tab.~\ref{table:ablation_1}, the smaller time window size negatively impacts the performance of the original AGRoL-MLP model ~\cite{du2023avatars}, which is trained on a longer window size to generate smooth motion sequences. In particular, Jitter increases significantly from $13.01$ to $22.70$ for the one trained with a smaller window size. To address this issue, we introduce a Memory Block that integrates past information to the hidden layers of backbone, while ensuring that the computational complexity remains acceptable. The second row represents the impact of Memory block using only the rotation branch. The results validate that Memory-block leads to significant improvements in both the MPJVE (from $35.77$ to $19.57$) and Jitter (from $22.70$ to $8.42$) metrics. Furthermore, the impact of multi-head predictor is reported in the third row of Tab.~\ref{table:ablation_1}. The results demonstrate that the performance for all metrics are also improved compared to the base model. Lastly, we combine Memory-Block and Multi-Head Predictor and report their performance in the last row. The results validate that combining both of our contributions achieves the state-of-the-art results.

\begin{table}[t!]
    \centering
    \resizebox{\columnwidth}{!}{%
    \begin{tabular}{lcccc}
        \hline
        \textbf{Method} & \textbf{MPJRE} \darr & \textbf{MPJPE} \darr & \textbf{MPJVE} \darr & \textbf{Jitter} \darr \\
        \hline
        \xmark Multi-Head , \xmark Memory-Block & $2.76$ & $4.13$ & $35.77$ & $22.70$ \\
        \xmark Multi-Head , \cmark Memory-Block & $2.53$ & $3.72$ & $19.57$ & $8.42$ \\
        \cmark Multi-Head , \xmark Memory-Block & $2.58$ & $3.43$ & $18.17$ & $7.79$ \\
        \cmark Multi-Head , \cmark Memory-Block & $2.57$ & $3.08$ & $15.02$ & $6.03$  \\
        \hline
    \end{tabular}}
    \vspace{-0.8em}
    \caption{Impact of Multi-Head Predictor and Memory Block. Results provide comparisons for the different configurations of our model.}
    \vspace{-0.8em}
    \label{table:ablation_1}
\end{table}


\begin{table}[t!]
    \centering
    \resizebox{\columnwidth}{!}{%
    \begin{tabular}{ccccc}
        \hline
        \textbf{\makecell{Automatic Loss \\ Weighting}} & \textbf{MPJRE} \darr & \textbf{MPJPE} \darr & \textbf{MPJVE} \darr & \textbf{Jitter} \darr \\
        \hline
        \xmark  & $2.60$ & $3.28$ & $17.40$ & $7.46$ \\
        \cmark  & $2.57$ & $3.08$ & $15.02$ & $6.03$ \\
        \hline
    \end{tabular}}
    \vspace{-0.8em}
    \caption{Impact of Loss Weighting Mechanism. The results show that the automatic loss weighting mechanism is important for the state-of-the-art performance.}
    \label{table:loss_weights}
    \vspace{-1.8em}
\end{table}

\noindent
\textbf{Impact of Loss Weighting Mechanism:}
Our model is trained using multiple losses, where each loss function serves a crucial purpose. However, these losses can be conflicting, which can hinder the training process. To address this issue, we leverage homoscedastic uncertainty to modulate the contributions of each loss. By doing so, we enable the model to adaptively learn the optimal loss weights during training, allowing it to balance the competing objectives and improve overall performance. Specifically, to demonstrate the impact of this weighting mechanism, we perform an additional ablation experiment. In this experiment, we employ manually selected loss weights, optimized through a grid search, for each loss in our objective function. From grid-search, we set \( \lambda_{\theta} = 1 \), \( \lambda_{rv} =30 \), \( \lambda_{p} = 0.5 \), \( \lambda_{fv} = 0.1\), and \( \lambda_h  = 0.1\). Table \ref{table:loss_weights} shows the results for this experiment. As demonstrated, the automatic loss weighting mechanism significantly improves the performance of our model across all metrics, particularly, a notable enhancement for the Jitter metric.

\vspace{-0.8em}

\section{Conclusions}
 In this paper, we introduce a novel method, \textit{Mem-MLP}, for generating 3D full-body motions from sparse inputs, achieving state-of-the-art accuracy while being sufficiently lightweight for real-time applications. In particular, our method leverages an MLP backbone with residual connections and introduces a novel component called Memory-Block that represents missing sensor inputs with trainable code-vectors. Ultimately, temporal consistency and smoothness are improved with the code-vectors and sparse signals from previous time instances. Furthermore, we formulate our solution as a multi-task learning problem, where losses from rotation and orientation predictions are jointly optimized. The experiments on the AMASS dataset demonstrate that the superiority of our method over SOTA methods in terms of model complexity and model accuracy.


{
    \small
    \bibliographystyle{ieeenat_fullname}
    \bibliography{main}
}

\newpage

\clearpage 
\appendix   






\newcommand{\nbf}[1]{{\noindent \textbf{#1.}}}

\def\wacvPaperID{181}
\def\confName{WACV}
\def\confYear{2026}

\frenchspacing



\maketitlesupplementary

In the ablation study, we also conduct additional experiments on memory block, multi-task predictor, number of MLP blocks,  sequence length, impact of VQ-VAE and comparison with LSTM and GRU alternatives. All methods are evaluated for Scenario-1 (S1).

\label{sec:ablation_study}
\label{sec:designchoices}

\begin{table}[t!]
    \centering
    \resizebox{\columnwidth}{!}{%
    \begin{tabular}{lcccc}
        \toprule
        \textbf{Method} & \textbf{MPJPE} ↓ & \textbf{Upper PE} ↓ & \textbf{Lower PE} ↓ & \textbf{Jitter} ↓ \\
        \midrule
        Rotation Branch & $3.36$ & $1.56$ & $5.97$ & $7.34$ \\
        Position Branch  & $3.07$ & $1.24$ & $5.72$ & $5.45$  \\
        Rotation + Position Branch  & $3.08$ & $1.25$ & $5.72$ & $6.03$  \\
        \bottomrule
    \end{tabular}}
    \caption{Experiments on inverse kinematics. We run an optimized inverse kinematics solution to correct the estimated angles (rotation branch) using 3D joint locations (position branch) as target points.}
    \label{table:ablation_ik}
\end{table}

\section{Architectural Design}
\label{arch_design_sup}
\noindent
\textbf{Inverse Kinematics:}
We observe that the estimated 3D joint positions generated by the second branch of our multi-head predictor are more accurate than those from the first branch. However, since the SMPL model cannot directly reconstruct 3D body pose representations from joint positions alone, we employ an inverse kinematics (IK) solver using the L-BFGS optimization method, achieving optimal results within 15 iterations. The IK solver optimizes joint rotations by initializing with the output of the rotation branch and targeting the more accurate 3D joint positions from the second branch. The corresponding results are presented in Table~\ref{table:ablation_ik}.

\begin{table}[t!]
    \centering
    \resizebox{\columnwidth}{!}{%
    \begin{tabular}{cccccc}
        \toprule
        \textbf{\#Layers (PB)} & \textbf{\#Layers (RB)} & \textbf{MPJRE} \darr & \textbf{MPJPE} \darr & \textbf{MPJVE} \darr & \textbf{Jitter} \darr \\
        \midrule
        $1$ & $1$  & $2.60$ & $3.14$ & $15.15$ & $6.04$ \\
        $2$ & $2$  & $2.57$ & $3.08$ & $15.02$ & $6.03$ \\
        $4$ & $4$  & $2.57$ & $3.11$ & $15.04$ & $6.41$ \\
        \bottomrule
    \end{tabular}}
    \caption{Experiments on layer depth of multi-task predictor. PB and RB denote position and rotation branches.}
    \label{table:heads}
\end{table}

\vspace{5pt}

\noindent
\textbf{Layer Depth of Multi-Task Predictor:}
In Table~\ref{table:heads} we evaluate the effect of different numbers of MLP blocks per branch. Our default configuration (2 layers) yields the best trade-off, especially reducing Jitter by 16.1\% over 1-layer and 13.9\% over 4-layer configurations.

\begin{table}[t!]
    \centering
    \resizebox{\columnwidth}{!}{%
    \begin{tabular}{lcccc}
        \toprule
        \textbf{Method} & \textbf{MPJRE} \darr & \textbf{MPJPE} \darr & \textbf{MPJVE} \darr & \textbf{Jitter} \darr \\
        \midrule
        MHP w/ MB & $2.59$ & $3.40$ & $17.80$ & $5.57$ \\
        MHP wo/ MB  & $2.57$ & $3.08$ & $15.08$ & $6.03$  \\
        \bottomrule
    \end{tabular}}
    \caption{Effect of Memory-Block (MB) integration into Multi-Head Predictor (MHP).}
    \label{table:ablation_memory}
\end{table}

\vspace{5pt}

\noindent
\textbf{Layer Depth of MLP Backbone:} We conduct experiments with different number of MLP blocks in the backbone of our architecture. As presented in \ref{table:ablation_mlpblocks}, our default configuration ($L=8$) presents better MPJRE and Jitter results, while ($L=10$), shows an improvement in MPJPE and MPJVE metrics, but increases the computational overhead in our model, which falls slightly under real-time performance.

\begin{table}[t!]
    \centering
    \resizebox{\columnwidth}{!}{%
    \begin{tabular}{cccccc}
        \toprule
        \textbf{MLP Blocks $L$} & \textbf{MPJRE} \darr & \textbf{MPJPE} \darr & \textbf{MPJVE} \darr & \textbf{Jitter} \darr & 
        \textbf{FLOPS} \darr\\
        \midrule
        8 & $2.57$ & $3.08$ & $15.02$ & $6.03$ & $0.25$G \\
        10  & $2.58$ & $3.07$ & $14.78$ & $6.19$ & $0.27$G\\
        12  & $2.59$ & $3.11$ & $14.88$ & $6.10$ & $0.30$G \\
        \bottomrule
    \end{tabular}}
    \caption{Effect of Number of MLP blocks $L$ on model predictive accuracy.}
    \label{table:ablation_mlpblocks}
\end{table}

\vspace{5pt}

\noindent
\textbf{Integration of Memory-Block to Multi-Head Predictor:}
It is observed that, Table~\ref{table:ablation_memory}, including MB in MHP improves temporal consistency (lower Jitter), although MPJRE and MPJPE are slightly worse. This shows that MB smooths the motion by leveraging past context.

\begin{table}[t!]
    \centering
    \resizebox{\columnwidth}{!}{%
    \begin{tabular}{ccccc}
        \toprule
        \textbf{Seq. Length $T$} & \textbf{MPJRE} \darr & \textbf{MPJPE} \darr & \textbf{MPJVE} \darr & \textbf{Jitter} \darr \\
        \midrule
        $11$  & $2.60$ & $3.27$ & $20.81$ & $15.05$ \\
        $21$  & $2.61$ & $3.19$ & $16.82$ & $8.22$ \\
        $41$  & $2.57$ & $3.08$ & $15.02$ & $6.03$ \\
        $60$  & \textbf{2.56} & \textbf{3.05} & \textbf{14.27} & \textbf{5.35} \\
        \bottomrule
    \end{tabular}}
    \caption{Impact of sequence length $T$ on model performance.}
    \label{table:ablation_c}
\end{table}

\vspace{5pt}

\noindent
\textbf{Impact of Sequence Length:}
Longer sequences improve performance (especially Jitter and MPJVE), but increase computation, Table~\ref{table:ablation_c}. This trade-off must be tuned depending on application constraints.

\begin{table}[t!]
    \centering
    \resizebox{\columnwidth}{!}{%
    \begin{tabular}{lcccc}
        \toprule
        \textbf{Method} & \textbf{MPJRE} \darr & \textbf{MPJPE} \darr & \textbf{MPJVE} \darr & \textbf{Jitter} \darr \\
        \midrule
        Mem-MLP w/o VQ-VAE & $2.60$ & $3.16$ & $15.35$ & $6.61$ \\
        Mem-MLP w/ VQ-VAE  & $2.57$ & $3.08$ & $15.08$ & $6.03$  \\
        \bottomrule
    \end{tabular}}
    \caption{Effect of VQ-VAE in Memory Block.}
    \label{table:ablation_vqvae}
\end{table}

\vspace{5pt}

\noindent
\textbf{Impact of VQ-VAE:}
To better understand the impact of code-vector $\mathcal{E}_l$ in the Memory Block of our architecture we conduct another experiment where we remove the code-vector from our Memory Block, and the feature $\mathbf{Z}_{m}$ is given by the following equation: $\mathbf{Z}_{m}=\mathbf{m} * \mathbf{Z}_{\theta}$. Table~\ref{table:ablation_vqvae} shows that the incorporation of code-vector $\mathcal{E}_l$ in our Memory Block enhances the predictive accuracy of our model in all metrics. In our experiments, we set the dimension of the code-vector to 64. 

\noindent
\textbf{VQ-VAE Details:}
Below we report extra details on the VQ-VAE model. The overall architecture is illustrated in Figure~1. We adopt a lightweight MLP-based design: each MLP block comprises layer normalization, a single linear layer and a SiLU activation; this block is used uniformly in both the encoder and decoder. The encoder depth is denoted $\mathbf{L_{enc}}$, with baseline value $\mathbf{L_{enc}}=4$, the decoder depth $\mathbf{L_{dec}}$, with baseline value $\mathbf{L_{dec}}=1$ and the latent dimensionality is $d_{z_s}=256$ of the model. The training is performed trying to minimize a loss that consists of two terms, the rotation reconstruction loss (Equation~\ref{rotation_loss}), and the commitment loss introduced in ~\cite{NIPS2017_7a98af17}, which encourages the encoder outputs to commit to specific codebook entries. The discrete codebook is denoted $\mathcal{C} = \{ \mathbf{e}_1, \dots, \mathbf{e}_K \} \subset \mathbb{R}^{d_{z_s}}$, where $K=64$ is the number of codebook entries in our best performing Mem-MLP models. This choice yields a compact yet expressive discrete representation suitable for downstream reconstruction and prior modeling. Mem-MLP pipeline is only using the pre-trained, frozen, encoder of a VQ-VAE model during training and inference. This design reduces training time while improving accuracy. We also experimented with joint training of the VQ-VAE and Mem-MLP, but observed substantial performance degradation. We train the VQ-VAE for 100 epochs using the Adam optimizer with batch size 256. The Adam hyperparameters are set to $\beta=(0.9, 0.99)$ and a weight decay of $10^{-4}$. The initial learning rate is $1 \times 10^{}-4$  and it is decreased by a factor of 0.2 at scheduled milestone epochs [20, 50, 70].
\begin{figure}
    \centering
    \includegraphics[width=0.8\linewidth]{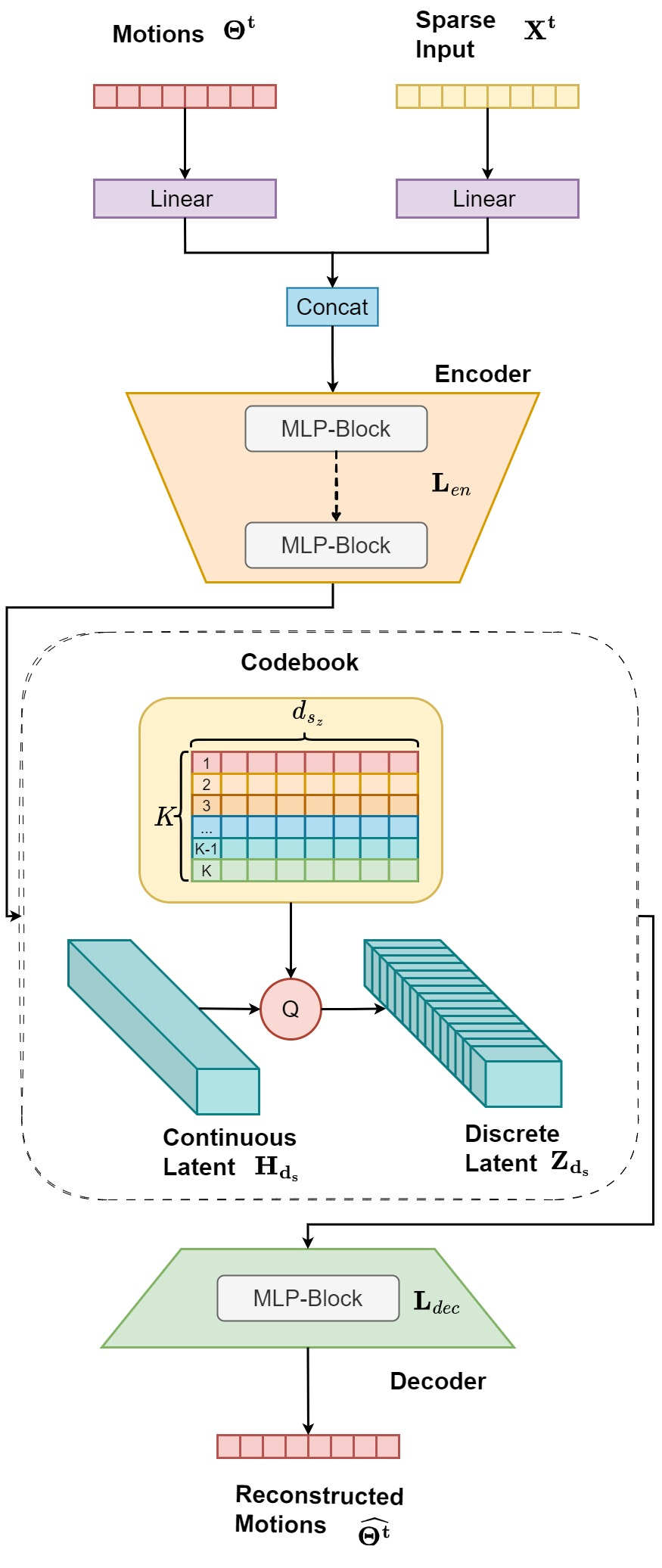}
    \caption{VQ-VAE architecture diagram}
    \label{fig:vqvae_arch}
\end{figure}

\noindent
\textbf{VQ-VAE Configuration Settings:} To quantify the sensitivity of downstream performance to the VQ-VAE configuration, we perform ablation experiments over codebook sizes $K \in {32, 64, 128}$ and the encoder depths $\mathbf{L_{enc}} \in{2, 4, 6}$. For each VQ-VAE variant we freeze the learned codebook and integrate the corresponding discrete priors into the final Mem-MLP model. More specifically, Table \ref{table:albation_lenc_vqvae} presents the results for varying encoder depths $\mathbf{L_{enc}}$, while keeping all other VQ-VAE hyper-parameters ($\mathbf{L_{dec}}$, $d_{z_s}$, and $K=64$) fixed. Each trained VQ-VAE is subsequently employed within Mem-MLP, where a new model is trained from scratch under the corresponding configuration. In parallel, Table \ref{table:albation_k_vqvae} reports the ablation results on the VQ-VAE codebook size $K$, with all remaining hyperparameters held constant.

\begin{table}[t!]
    \centering
    \resizebox{\columnwidth}{!}{%
    \begin{tabular}{lccccc}
        \toprule
        \textbf{Model} & $\mathbf{L_{enc}}$ & \textbf{MPJPE} ↓ & \textbf{Upper PE} ↓ & \textbf{Lower PE} ↓ & \textbf{Jitter} ↓ \\
        \midrule
        Mem-MLP & 2 & $2.59$ & $3.11$ & $15.13$ & $6.08$ \\
        Mem-MLP  & 4 & $2.57$ & $3.08$ & $15.02$ & $6.03$  \\
        Mem-MLP  & 6 & $2.58$ & $3.15$ & $15.34$ & $6.21$  \\
        \bottomrule
    \end{tabular}}
    \caption{Experiments on different sizes of $\mathbf{L_{enc}}$ and their impact on Mem-MLP.}
    \label{table:albation_lenc_vqvae}
\end{table}

\begin{table}[t!]
    \centering
    \resizebox{\columnwidth}{!}{%
    \begin{tabular}{lccccc}
        \toprule
        \textbf{Model} & $\mathbf{K}$ & \textbf{MPJPE} ↓ & \textbf{Upper PE} ↓ & \textbf{Lower PE} ↓ & \textbf{Jitter} \\
        \midrule
        Mem-MLP & $32$ & $2.57$ & $3.13$ & $15.30$ & $6.14$ \\
        Mem-MLP  & $64$ & $2.57$ & $3.08$ & $15.02$ & $6.03$  \\
        Mem-MLP  & $128$ & $2.62$ & $3.20$ & $15.55$ & $6.52$  \\
        \bottomrule
    \end{tabular}}
    \caption{Experiments on different sizes of $K$, and their impact on Mem-MLP.}
    \label{table:albation_k_vqvae}
\end{table}

\begin{figure}
    \centering
    \includegraphics[width=0.5\linewidth]{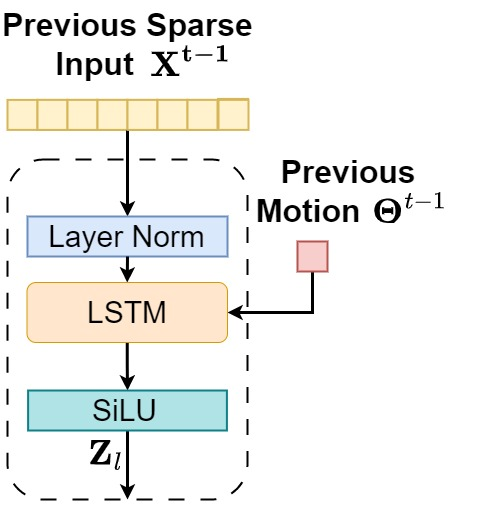}
    \caption{LSTM-based memory block diagramm. }
    \label{fig:lstm_block_vis}
\end{figure}

\noindent
\textbf{Comparison with LSTM \& GRU Memory Blocks:} To further assess the effectiveness of our approach, we implemented recurrent memory variants by replacing the memory block with LSTM- or GRU-based layers. The architecture is extended with a recurrent memory block that applies Layer Normalization to the input sparse features, followed by either an LSTM or GRU layer, and a SiLU activation function.
During training, the recurrent units are initialized differently depending on the configuration: 
(a) LSTM-based memory block: The hidden state is initialized using the final time step of the sparse motion features, thereby incorporating prior motion context, while the cell state is zero-initialized.
(b) GRU-based memory block: The hidden state alone is initialized from the final time step of the sparse motion features. 
Figure \ref{fig:lstm_block_vis} provides visual explanation on the LSTM-based memory block structure.
In evaluation mode, both LSTM and GRU blocks are run without explicit initialization, relying instead on internally propagated states. Table \ref{table:ablation_lstm} reports the results, where our method consistently outperforms both RNN-based memory configurations. While the Memory-Block superficially resembles recurrent architectures because it reuses temporal context, its design differs in two critical aspects: (i) it avoids the sequential gating of LSTMs/GRUs, enabling higher parallelism and (ii) it fuses two complementary sources of information, current sparse inputs and VQ-VAE code-vectors (auto-regressively providing full-pose priors), yielding richer context than standard recurrence.

\textbf{Placement of Memory Blocks in Backbone:}
To further assess the impact of memory blocks on the performance of our model, we conduct an ablation study focusing on their placement within the backbone. Specifically, in our model, memory blocks are incorporated in some of the layers of backbone for computational efficiency. In particular, we introduce Memory-blocks in $2^{\text{nd}}$, $4^{\text{th}}$, $6^{\text{th}}$, and/or $8^{\text{th}}$ layers of the backbone.  The results presented in Table \ref{table:memblocks} illustrate the effect of progressively adding memory blocks to the backbone via different layers. From the results, we observe a constant improvement in performance as more memory block outputs are introduced into deeper layers of the backbone. Notably, incorporating memory blocks at all four layers obtains the best scores for MPJRE, MPJPE, and MPJVE metrics. However, a slightly lower Jitter score is obtained when memory block is introduced for the $8^{\text{th}}$ layer.

\begin{table}[t!]
    \centering
    \resizebox{\columnwidth}{!}{%
        \begin{tabular}{ccccc}
            \toprule
            \textbf{\makecell{Memory Block Layer(s) \\ in Backbone}}  & \textbf{MPJRE} \darr & \textbf{MPJPE} \darr & \textbf{MPJVE} \darr & \textbf{Jitter} \darr \\
            
            \midrule
            $2$ & $2.63$ & $3.17$ & $15.12$ &$6.82$ \\
            $2, 4$ & $ 2.63$ & $3.20$ & $15.57$ & $6.66$ \\
            $2, 4, 6$ &  $2.60$ & $3.14$ & $15.04$ & $6.37$ \\
            $2, 4, 6, 8$ & $\mathbf{2.57}$ & $\mathbf{3.08}$ & $\mathbf{15.02}$ & $\mathbf{6.03}$ \\
            \bottomrule
   \end{tabular}} 
    \caption{Impact of placement of memory blocks in backbone. The performance for all metrics improves when memory blocks are incorporated in the deeper layers of our backbone. The best results are in \textbf{bold}.}
    \label{table:memblocks}
\end{table}

\begin{table}[t!]
    \centering
    \resizebox{\columnwidth}{!}{%
    \begin{tabular}{lccccc}
        \toprule
        \textbf{Method} & \textbf{MPJRE} \darr & \textbf{MPJPE} \darr & \textbf{MPJVE} \darr & \textbf{Jitter} \darr & \textbf{FLOPs(G)} \darr \\
        \midrule
        GRU-MB & $2.63$ & $3.50$ & $19.11$ & $8.77$ & $0.24$\\
        LSTM-MB  & $2.59$ & $3.29$ & $17.40$ & $7.72$ & $0.25$ \\
        \textit{Ours}  & $2.57$ & $3.08$ & $15.02$ & $6.03$ & $0.25$ \\
        \bottomrule
    \end{tabular}}
    \caption{Assessment of LSTM and GRU as Alternatives to the Existing Memory Block.}
    \label{table:ablation_lstm}
\end{table}

\section{Model Size and Complexity}
\label{sec:speed_section}

\begin{table}[t!]
    \centering
    \resizebox{\columnwidth}{!}{%
    \begin{tabular}{c|ccc}
        \toprule
        \textbf{Method} & \textbf{FLOPs} \darr & \textbf{Size (MB)} \darr & \textbf{\#Params (M)} \darr \\
        \midrule
        AvatarPoser~\cite{jiang2022avatarposer} & $0.33$G & $15.73$ & $4.12$ \\ 
        AGRoL-MLP~\cite{du2023avatars} & $0.88$G & $\mathbf{14.25}$ & $\mathbf{3.74}$ \\
        AGRoL-Diffusion~\cite{du2023avatars} & $1.00$G & $28.54$ & $7.48$ \\
        AvatarJLM~\cite{zheng2023realistic} & $4.64$G & $243.41$ & $63.81$ \\
        BoDiffusion~\cite{castillo2023bodiffusion} & $0.46$G & $83.70$ & $21.94$ \\
        SAGE-Net~\cite{feng2024stratified} & $4.10$G & $458.79$ & $120.27$ \\
        EgoPoser~\cite{jiang2024egoposer} & $0.33$G & $15.77$ & $4.12$ \\
        MMD~\cite{dong2024realistic} & $7.98$G & $853.47$ & $101.72$ \\
        MANIKIN-S~\cite{jiang2024manikin} & $0.33$G & $-$ & $4.12$ \\
        MANIKIN-L~\cite{jiang2024manikin} & $4.64$G & $-$ & $63.8$ \\
        MANIKIN-LN~\cite{jiang2024manikin} & $4.64$G & $-$ & $63.8$ \\
        HDM-Poser~\cite{Dai_2024_CVPR} & $0.38$G & $66.55$ & $9.55$ \\
        \midrule
        Mem-MLP-41 (Ours) & \textbf{0.25}G & \underline{15.38} & \underline{4.10} \\
        Mem-MLP-60 (Ours) & 0.38G & 16.08 & 4.59 \\
        \bottomrule
    \end{tabular}}
    \caption{Model size and complexity (FLOPs, parameters, and memory).}
    \label{tab:size_flops_params}
\end{table}

\vspace{5pt}

\noindent
\textbf{Model Size and Model Complexity:}
It is crucial to consider the model size and model complexity (i.e.,  number of FLOPs) to better assess the computational efficiency and memory footprint. The results in Table \ref{tab:size_flops_params} highlight the trade-offs between these factors across state-of-the-art methods. Notably, \textit{Mem-MLP} demonstrates a strong balance across all three metrics. With a model size of 15.38 MB, it is the second most compact among the others, slightly higher than AGRoL-MLP \cite{du2023avatars}. This compactness positively affects the architecture, which minimizes storage requirements. Additionally, \textit{Mem-MLP} achieves the lowest FLOP (0.25G), outperforming all other methods, including AvatarPoser \cite{jiang2022avatarposer}, AGRoL-MLP \cite{du2023avatars}, and MANIKIN-S \cite{jiang2024manikin}, while being significantly more efficient than diffusion-based models such as AGRoL-Diffusion \cite{du2023avatars}, BoDiffusion \cite{castillo2023bodiffusion}, and MMD \cite{dong2024realistic}. 

Regarding the number of parameters, \textit{Mem-MLP} (4.10M) is the second lowest after AGRoL-MLP \cite{du2023avatars} (3.74M) and is notably lighter than diffusion-based models, such as MMD \cite{dong2024realistic} (101.72M) BoDiffusion \cite{castillo2023bodiffusion} (21.94M) and SAGE-Net \cite{feng2024stratified} (120.27M). Additionally, MANIKIN \cite{jiang2024manikin} achieves competing predictive accuracy with computational cost. Overall, these results indicate that our \textit{Mem-MLP} model achieves the optimal trade-off between efficiency and model complexity, making it well-suited for resource-constrained environments. Its reduced FLOPs and its compact size suggest that it can operate efficiently in real-time applications with minimal computational overhead.

\begin{table}[t!]
    \centering
    \begin{tabular}{c|cc}
        \toprule
        \textbf{Method} & \textbf{CPU (ms)} \darr & \textbf{CPU (FPS)} \uparr \\
        \midrule
        AvatarPoser~\cite{jiang2022avatarposer} & \textbf{6.0} & \textbf{72.0} \\ 
        AGRoL-MLP~\cite{du2023avatars} & $30.1$ & $26.0$ \\
        AGRoL-Diffusion~\cite{du2023avatars} & $290.5$ & $3.5$ \\
        Mem-MLP-41 (Ours) & \underline{6.8} & \textbf{72.0} \\
        Mem-MLP-60 (Ours) & $8.5$ & $65.0$ \\
        \bottomrule
    \end{tabular}
    \caption{On-device inference time (CPU). Lower is better for time (ms), higher is better for FPS.}
    \label{tab:inference_time}
\end{table}

\noindent
\textbf{On-Device Inference Speed:} We deployed a subset of methods (i.e., AvatarPoser \cite{jiang2022avatarposer}, AGRoL-MLP \cite{du2023avatars}, AGRoL-diffusion \cite{du2023avatars}, and ours) on the Meta Quest 3 to evaluate their on-device performance, Table~\ref{tab:inference_time}. This allows us to illustrate how each method can operate in real-world, resource-constrained environments. As previously noted, \textit{Mem-MLP-41} and AvatarPoser \cite{jiang2022avatarposer} achieve real-time performance, running at $6.8 ms$ and $6.0 ms$ on-device, respectively. Moreover, for \textit{Mem-MLP-60}, we observe that its execution time slightly increases to $8.5 ms$, maintaining near real-time performance. Nevertheless, this model remains a viable option for the applications where slightly higher latency is acceptable, when the methods are integrated with frame-interpolation techniques or a better HMD hardware with improved processing capabilities. In the deployment stage, all methods are converted to ONNX models and executed on a single CPU.

\section{Qualitative Results}
\begin{figure}
\begin{minipage}[b]{0.95\textwidth}
\centering

\begin{tabular}{cc}

     \includegraphics[width=0.95\textwidth]{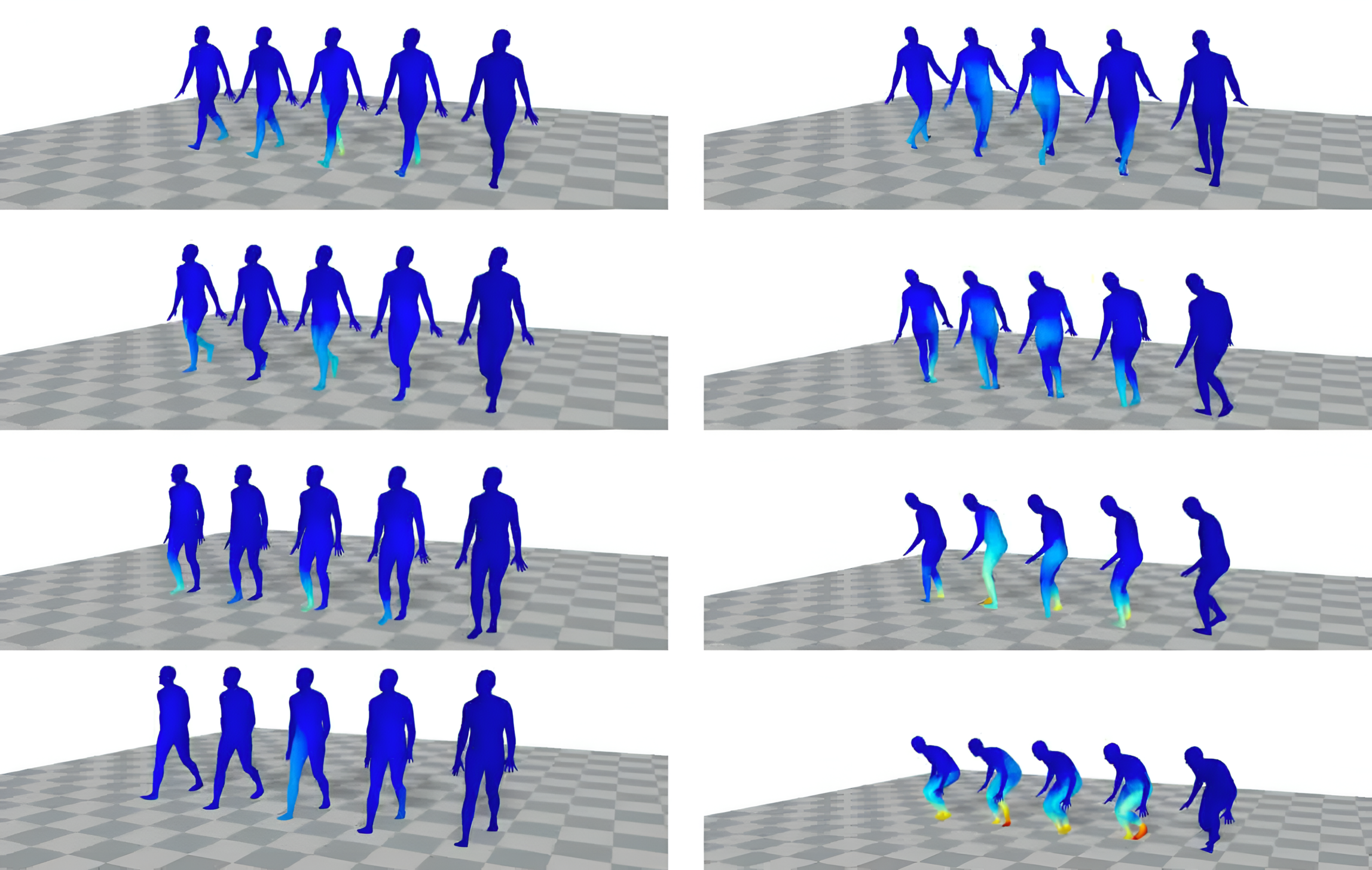}\hspace{-0.1em}  & 
     \includegraphics[width=0.04\textwidth]{Results/HeatMap_ErrorRange.png} \\
     \includegraphics[width=0.95\textwidth]{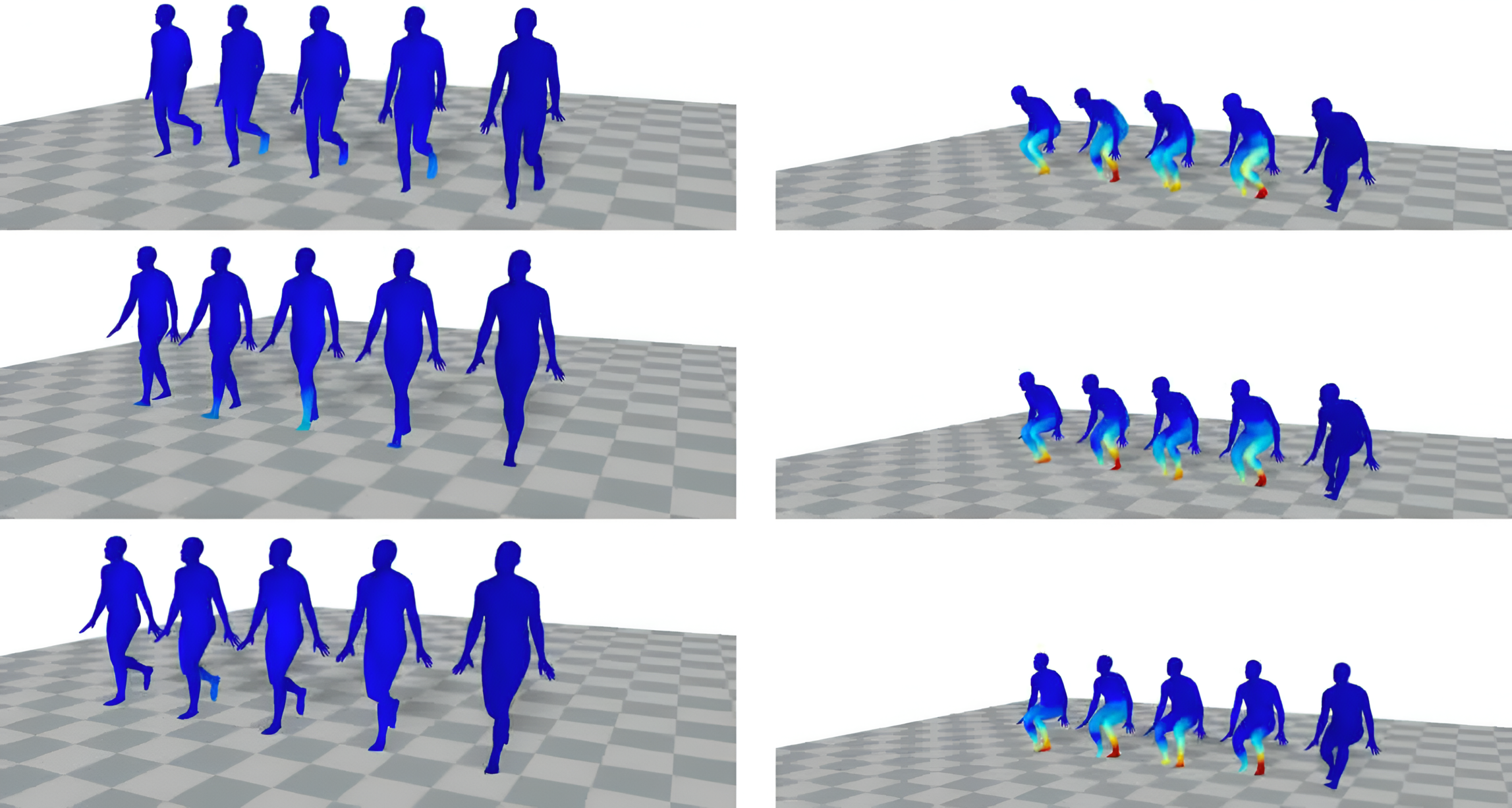}\hspace{-0.1em} & \\

\end{tabular}

\caption{Results on normal walk and sitting motions. Heat-map errors are plotted for different methods. From left to right: AvatarJLM \cite{zheng2023realistic}, AGRoL-Diffusion \cite{du2023avatars}, SAGE-Net \cite{feng2024stratified}, ours (\textit{Mem-MLP-41}) and the ground-truth.}
\label{Fig:poses_1}
\end{minipage}
\end{figure}

\begin{figure}
\begin{minipage}[b]{0.95\textwidth}
\centering
\begin{tabular}{cc}

\includegraphics[width=0.95\textwidth]{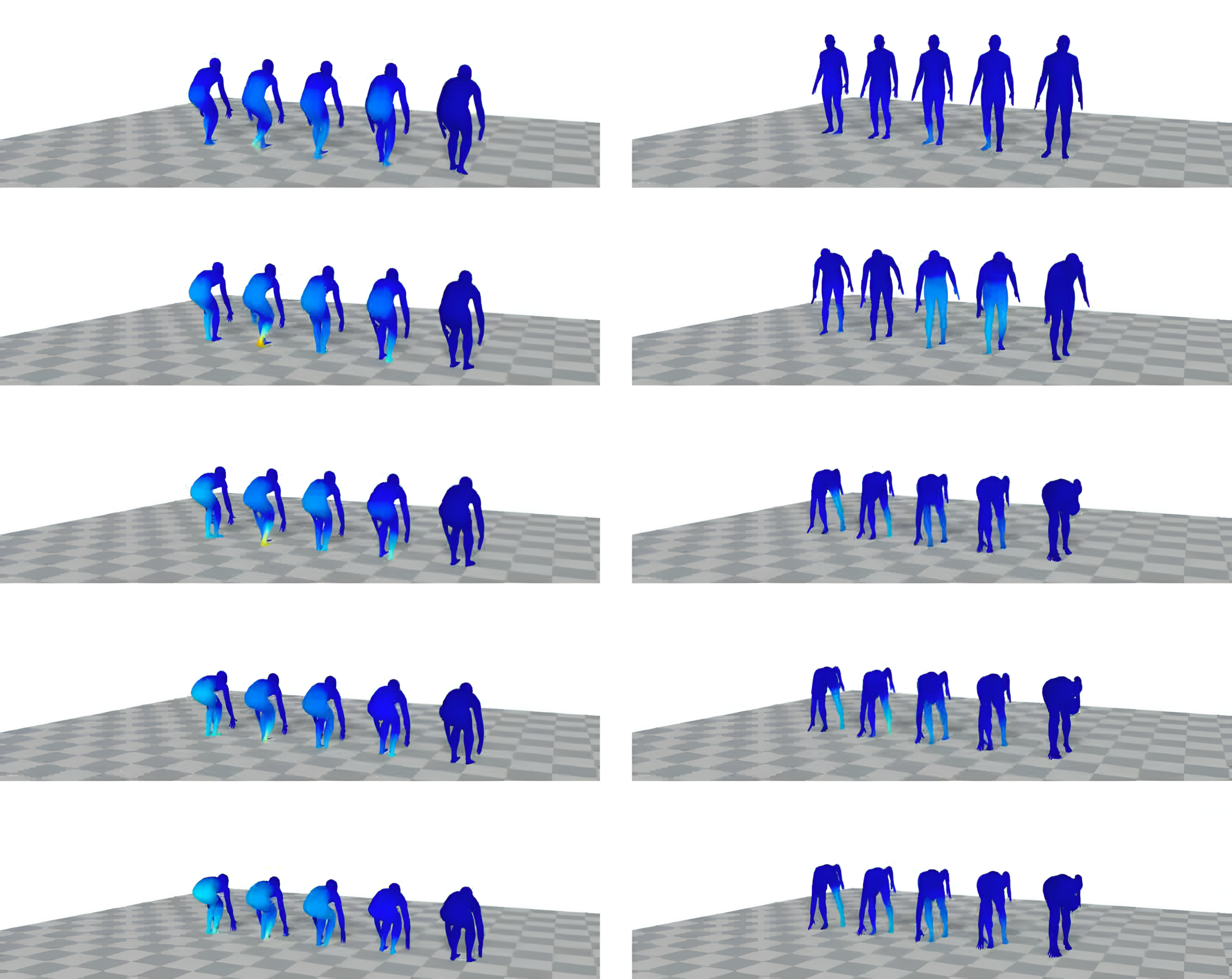}\hspace{-0.1em}  & 
\includegraphics[width=0.04\textwidth]{Results/HeatMap_ErrorRange.png} \\

\end{tabular}

\caption{Results on lifting and touching ground motions. Heat-map errors are plotted for different methods. From left to right: AvatarJLM \cite{zheng2023realistic}, AGRoL-Diffusion \cite{du2023avatars}, SAGE-Net \cite{feng2024stratified}, ours (\textit{Mem-MLP-41}) and the ground-truth.}
\label{Fig:poses_2}
\end{minipage}
\end{figure}

\begin{figure}
\begin{minipage}[b]{0.95\textwidth}
\centering
\begin{tabular}{cc}

\includegraphics[width=0.95\textwidth]{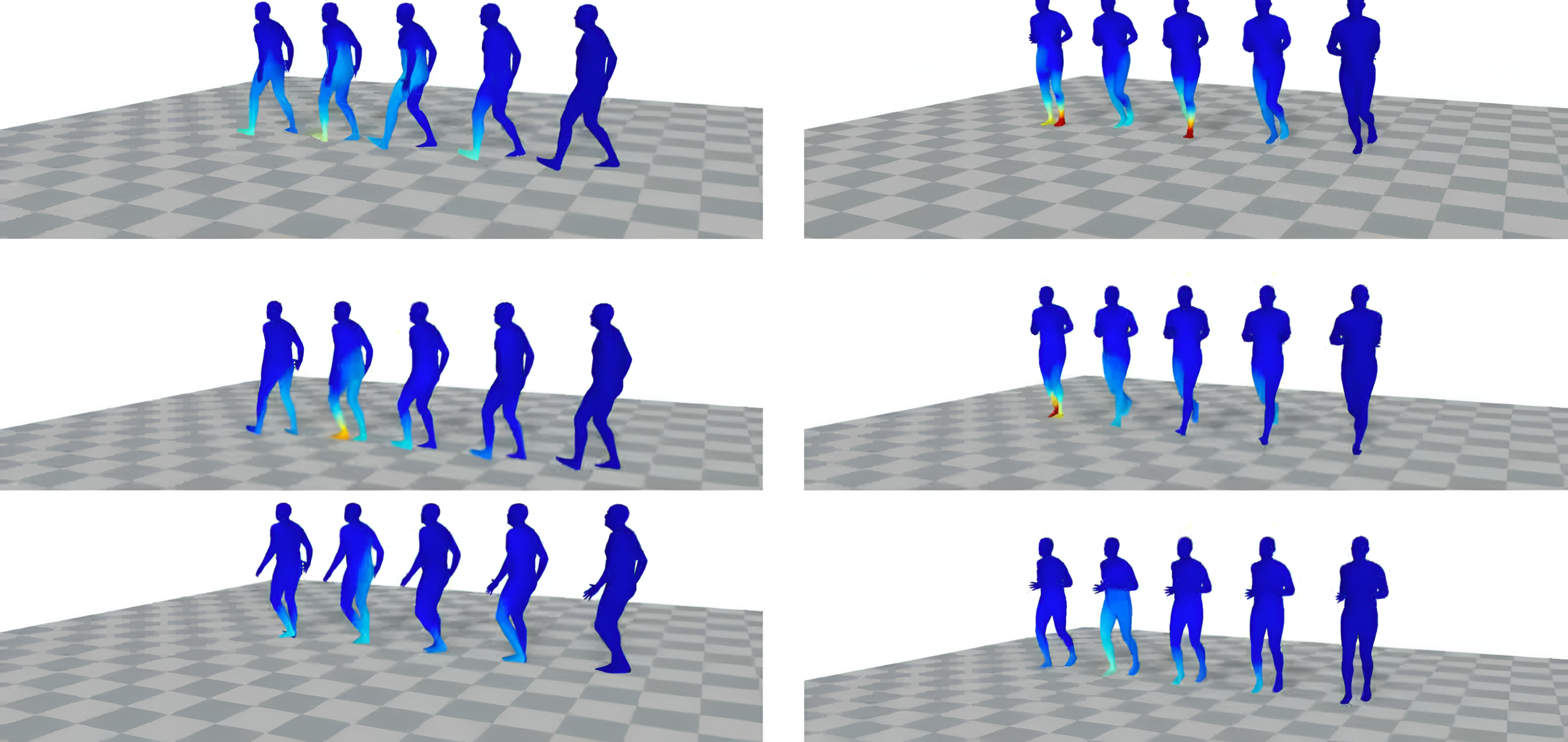}\hspace{-0.1em}  & 
\includegraphics[width=0.04\textwidth]{Results/HeatMap_ErrorRange.png} \\
\includegraphics[width=0.95\textwidth]{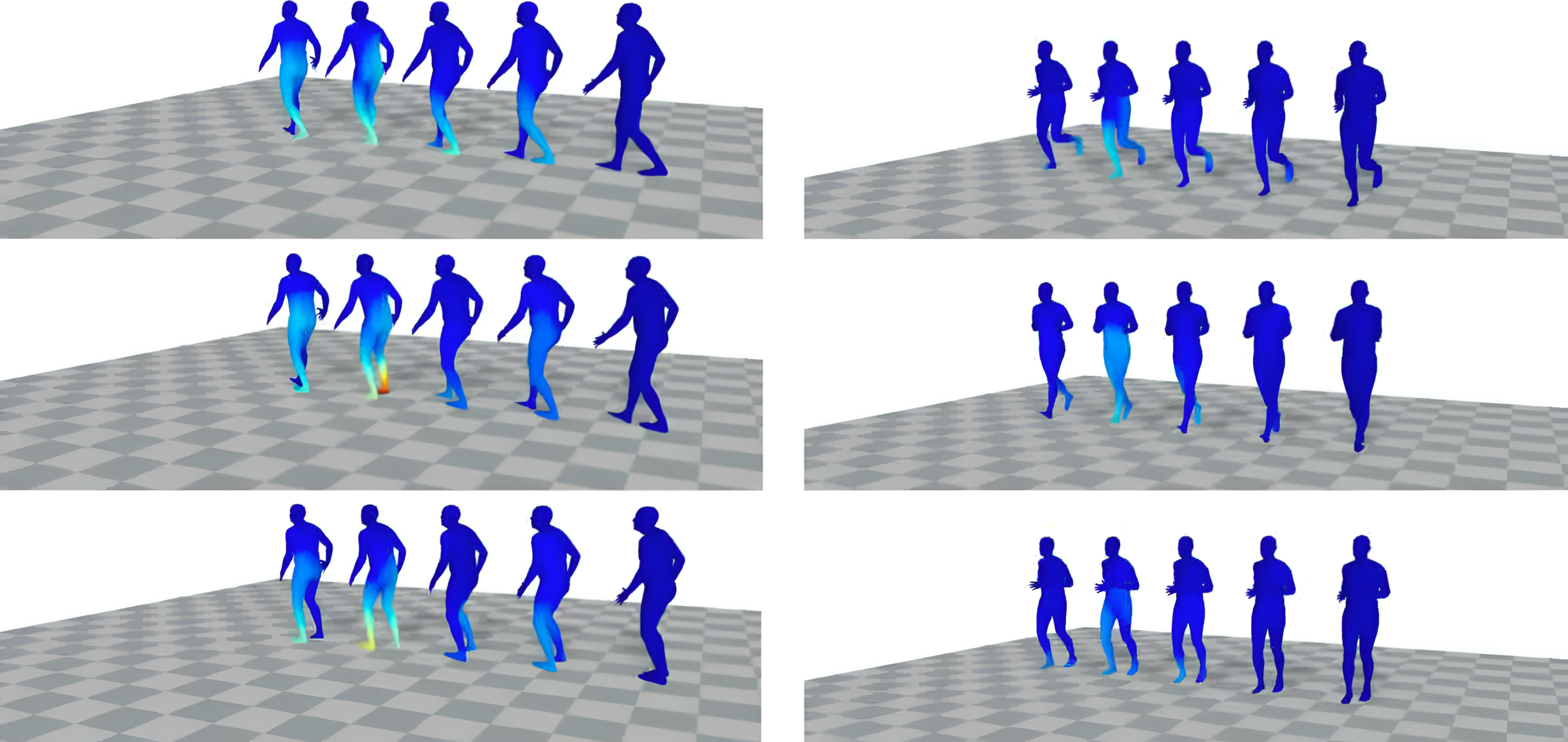}\hspace{-0.1em}  & \\

\end{tabular}

\caption{Results on backward walk and normal running motions.  Heat-map errors are plotted for different methods. From left to right: AvatarJLM \cite{zheng2023realistic}, AGRoL-Diffusion \cite{du2023avatars}, SAGE-Net \cite{feng2024stratified}, ours (\textit{Mem-MLP-41}) and the ground-truth.}
\label{Fig:poses_6}
\end{minipage}
\end{figure}

\begin{figure}
\begin{minipage}[b]{0.95\textwidth}
\centering
\begin{tabular}{cc}
 
\includegraphics[width=0.95\textwidth]{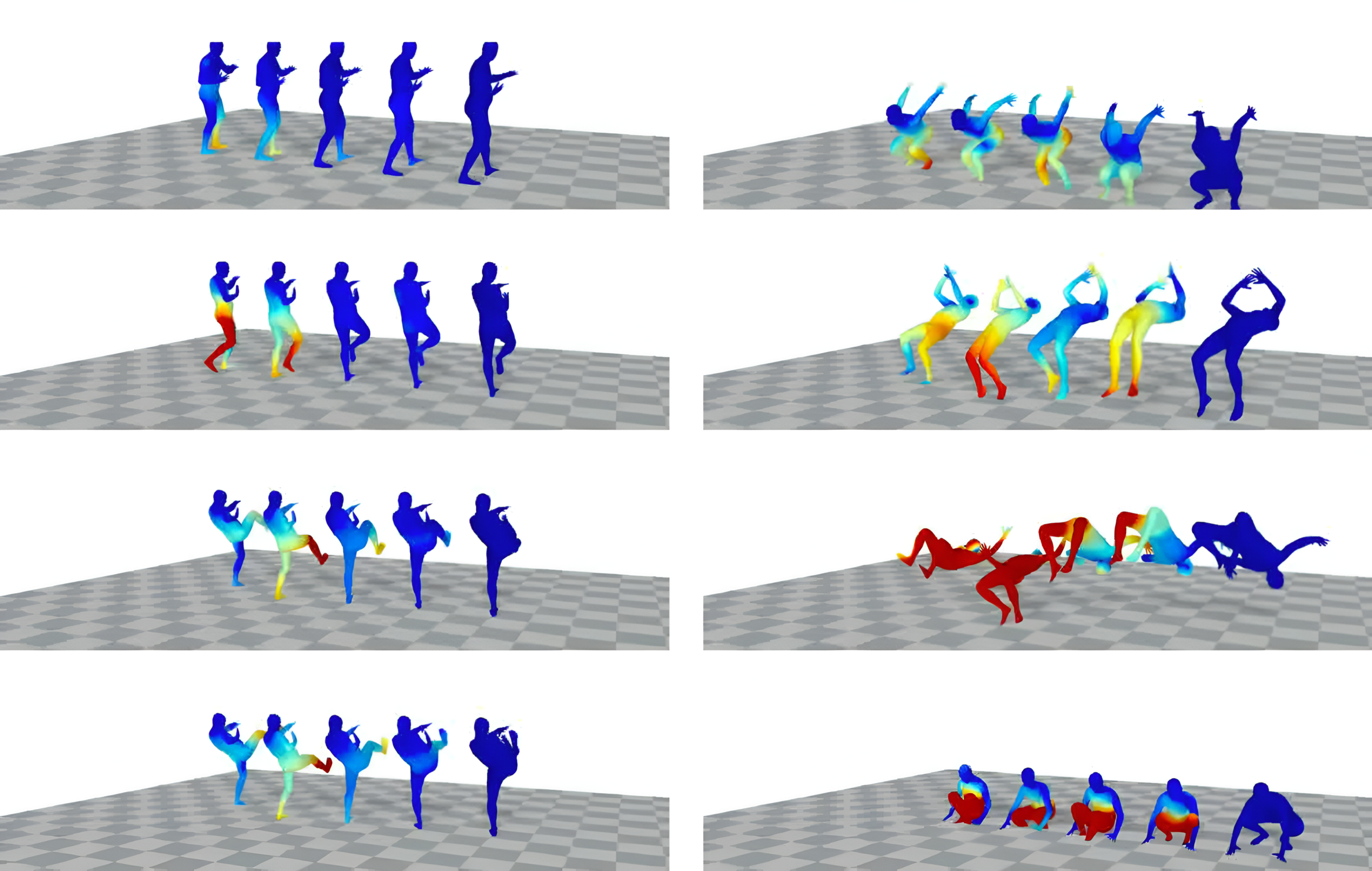}\hspace{-0.1em}  & 
\includegraphics[width=0.04\textwidth]{Results/HeatMap_ErrorRange.png} \\

\end{tabular}

\caption{Results on kicking and back-flip ground motions. Heat-map errors are plotted for different methods. From left to right: AvatarJLM \cite{zheng2023realistic}, AGRoL-Diffusion \cite{du2023avatars}, SAGE-Net \cite{feng2024stratified}, ours (\textit{Mem-MLP-41}) and the ground-truth.}
\label{Fig:poses_4}
\end{minipage}
\end{figure}

\begin{figure}
\begin{minipage}[b]{0.95\textwidth}
\centering
\begin{tabular}{cc}

\includegraphics[width=0.95\textwidth]{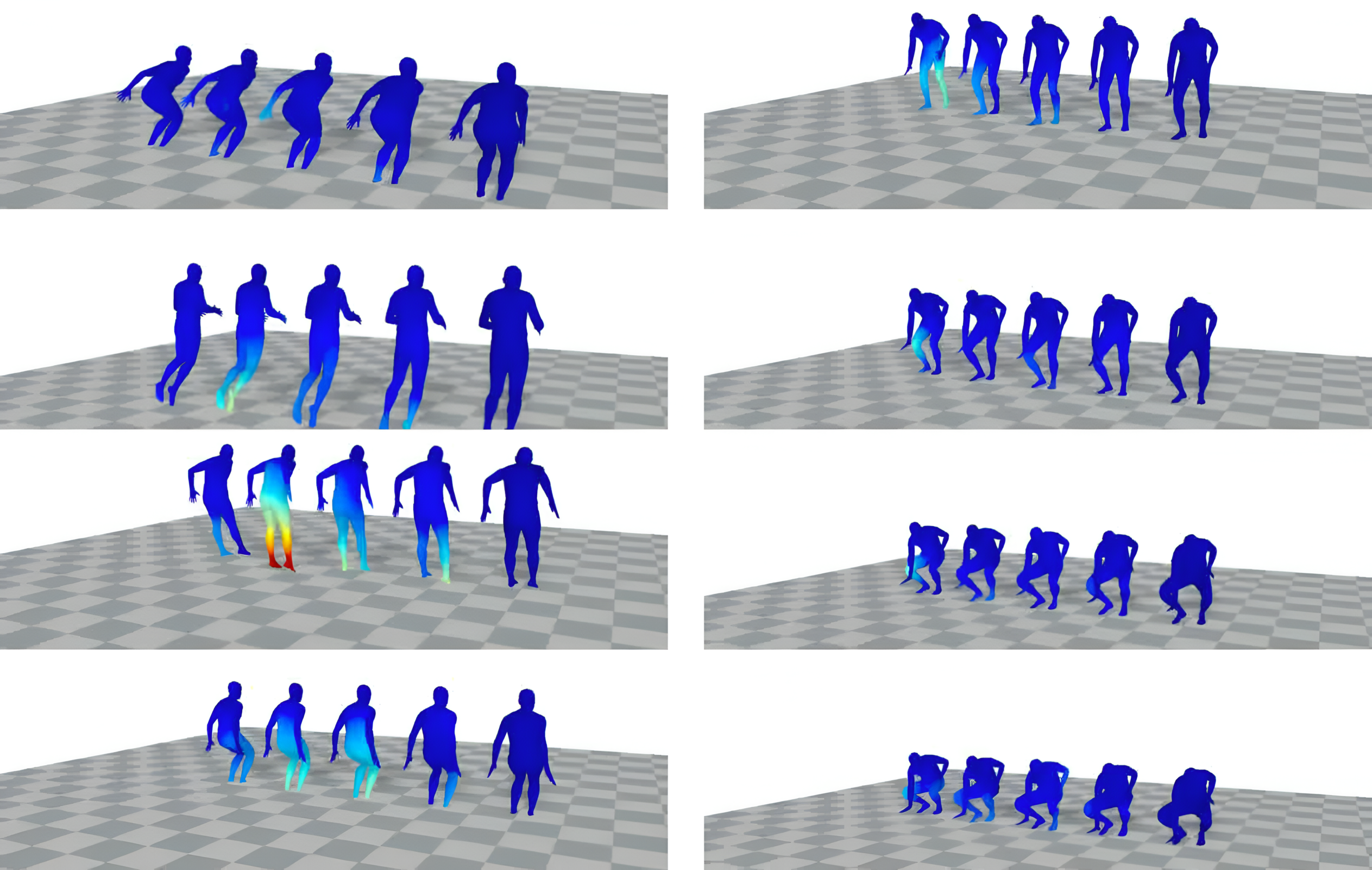}\hspace{-0.1em}  & 
\includegraphics[width=0.04\textwidth]{Results/HeatMap_ErrorRange.png} \\

\end{tabular}

\caption{Results on jumping and squating motions.  Heat-map errors are plotted for different methods. From left to right: AvatarJLM \cite{zheng2023realistic}, AGRoL-Diffusion \cite{du2023avatars}, SAGE-Net \cite{feng2024stratified}, ours (\textit{Mem-MLP-41}) and the ground-truth.}
\label{Fig:poses_5}
\end{minipage}
\end{figure}

We visualize the generation results of different methods for various motion types. The results are illustrated in Figs. ~\ref{Fig:poses_1}, ~\ref{Fig:poses_2}, ~\ref{Fig:poses_6}, 
~\ref{Fig:poses_4},  ~\ref{Fig:poses_5}. In particular, we report heat-map errors that ease the visualization errors for the different parts of human-body.

\onecolumn

\end{document}